\documentclass[10pt,twocolumn,letterpaper]{article}

\usepackage{cvpr}              % To produce the CAMERA-READY version
\usepackage{caption}
\usepackage{bbm}
\usepackage{arydshln}
\usepackage[accsupp]{axessibility}
\usepackage{listings}
\usepackage{placeins}
\usepackage{nicefrac}
\usepackage{xurl}
\usepackage{enumitem}
\usepackage{tabularx}
\usepackage{nicefrac}
\usepackage{makecell}
\usepackage{xspace}
\usepackage{graphbox}
\usepackage{pifont}
\usepackage{colortbl}
\usepackage{multirow}
\usepackage{diagbox}
\usepackage{array}
\usepackage{wrapfig}
\usepackage{subcaption}
\usepackage{footmisc}
\usepackage{color}
\usepackage{stfloats}
\usepackage{slashbox}
\usepackage{diagbox}
\usepackage{float}
\usepackage{mathrsfs}
\usepackage{apptools}
\usepackage{tabularray}
\usepackage{arydshln}
\usepackage{graphicx}
\usepackage{colortbl}
\usepackage{adjustbox}
\usepackage{tikz}
\usepackage{rotating}

\usepackage[pagebackref,breaklinks,colorlinks,allcolors=cvprblue]{hyperref}

\usepackage[dvipsnames]{xcolor}
\usepackage{tcolorbox}   % For colored boxes

\newcolumntype{C}[1]{>{\centering\arraybackslash}p{#1}}
\newcolumntype{L}[1]{>{\raggedright\arraybackslash}p{#1}}
\newcolumntype{R}[1]{>{\raggedleft\arraybackslash}p{#1}}
\newlength\newl
\newlength\newlc

\newlength\colwidth
\newlength\figwidth

\newcommand{\cmark}{\ding{51}}%
\newcommand{\xmark}{\ding{55}}%

\newcommand{\cleanclip}{CleanCLIP\xspace}
\newcommand{\clip}{CLIP\xspace}
\newcommand{\roclip}{RoCLIP\xspace}
\newcommand{\safeclip}{SafeCLIP\xspace}
\newcommand{\badnet}{BadNet\xspace}
\newcommand{\blended}{Blended\xspace}
\newcommand{\wanet}{WaNet\xspace}
\newcommand{\llava}{LLaVA\xspace}
\newcommand{\imnet}{ImageNet\xspace}
\newcommand{\ours}{PAR\xspace}%Anti-Back\xspace}
\newcommand{\badclip}{BadCLIP\xspace}
\newcommand{\ccm}{CC3M\xspace}
\newcommand{\synth}{SynthCLIP\xspace}
\newcommand{\vit}{ViT\xspace}

\definecolor{cvprblue}{rgb}{0.21,0.49,0.74}
\definecolor{bluey}{rgb}{0.95,0.95,0.7}
\definecolor{NewGray}{rgb}{0.85,0.85,0.9}
\definecolor{lightblue}{rgb}{0.7,0.9,1.0}
\definecolor{lightpink}{rgb}{1.0,0.8,0.88}
\definecolor{newpink}{rgb}{0.9,0.2,0.55}
\definecolor{lightgreen}{rgb}{0.9,1.0,0.9}

\newcommand{\addi}[1]{{\color{black}#1}}
\newcommand{\mh}[1]{{\color{black}#1}}

\newcommand\blfootnote[1]{%
  \begingroup
  \renewcommand\thefootnote{}\footnote{#1}%
  \addtocounter{footnote}{-1}%
  \endgroup
}
\usepackage[cal=boondox,scr=boondoxo]{mathalfa}

\newcommand{\inner}[1]{\left\langle#1\right\rangle}

\def\K{\mathcal{K}}

\def\L{\mathcal{L}}

\makeatletter
\def\on@dot@space{\ifx\@let@token.\else.\@\fi\xspace}
\makeatother

\def\eg{\emph{e.g}\onedot}

\newcommand*{\myparagraph}[1]{\par\vspace{\baselineskip}\noindent\textbf{#1}}
\def\confName{CVPR}
\def\confYear{2026}

\title{
Perturb and Recover: Fine-tuning for Effective Backdoor Removal from CLIP
}

\author{Naman Deep Singh$^{1}$ \qquad Francesco Croce$^{2}$ \qquad Matthias Hein$^{1}$ \vspace{0.3em} \\
{\normalsize $^1$University of Tübingen \& Tübingen AI Center, Germany} \quad {\normalsize $^2$EPFL, Switzerland}
}

\begin{document}

\maketitle
\blfootnote{Correspondence: \texttt{naman-deep.singh@uni-tuebingen.de}}

\begin{abstract}
    Vision-Language models like CLIP have been shown to be highly effective at linking visual perception and natural language understanding, enabling sophisticated image-text capabilities, including strong retrieval and zero-shot classification performance. Their widespread use, as well as the fact that CLIP models are trained on image-text pairs from the web, make them both a worthwhile and relatively easy target for backdoor attacks. As training foundational models, such as CLIP, from scratch is very expensive, this paper focuses on cleaning potentially poisoned models via fine-tuning. We first show that existing cleaning techniques are not effective against simple structured triggers used in Blended or BadNet backdoor attacks, exposing a critical vulnerability for potential real-world deployment of these models. 
Then, we introduce PAR, Perturb and Recover, a surprisingly simple yet effective mechanism to remove backdoors from CLIP models. Through extensive experiments across different encoders and types of backdoor attacks, we show that PAR achieves high backdoor removal rate while preserving good standard performance. Finally, we illustrate that our approach is effective even only with synthetic text-image pairs, i.e. without access to real training data. The code and models are available on \href{https://github.com/nmndeep/PerturbAndRecover}{GitHub}. 
% Vision-Language Models like CLIP have been effective in bridging the gap between visual perception and natural language understanding, enabling sophisticated image/text capabilities. Wide-spreaad use of these models also puts under threat their robustness against backdoor attacks. While some defenses have been proposed for cleaning backdoored CLIP model, we show they can be easily circumvented by adding simple structured triggers in known attacks. Through a series of such triggers, we show how attackers can easily bypass defenses that rely heavily on data augmentations. This discovery fundamentally challenges the reliability of existing defense strategies and exposes critical weaknesses in systems currently deployed in real-world applications. Further, to tackle such attacks, we introduce \ours, a surprisingly simple yet effective cleaning mechanism for pre-training backdoored multi-modal models. By leveraging distillation to unlearn the spurious correlations learnt during poisoning, \ours enables models to ``forget" backdoor triggers while preserving their core capabilities. Through extensive experiments across difference encoders and against different types of backdoor attacks, we show \ours achieves high backdoor removal rate.
\end{abstract}

\section{Introduction}
\label{sec:intro}

Multi-modal models like CLIP~\cite{radford2021learning}, ALIGN~\cite{jia2021scaling} and BLIP-2~\cite{li2023blip} are trained on large datasets to map multiple modalities in a joint-embedding space.
% \fra{next two sentences say almost the same thing} \clip is very effective in applications where multiple modalities can be used as input, making them pervasive in a lot of applications. 
These models are often used for a variety of tasks,  including image/text retrieval, zero-shot classification, image captioning, question answering etc.
Moreover, CLIP's vision encoder forms an essential part of large vision-language models (LVLMs) like \llava~\cite{liu2024improved}, VILA~\cite{lin2024vila} and CogVLM~\cite{wang2023cogvlm} where it is used in combination with large language models (LLMs) like Vicuna~\cite{vicuna2023}, Qwen~\cite{qwen1.5} or Llama 3~\cite{dubey2024llama}. 

\begin{figure}[t]
\small \centering
\begin{tikzpicture}

      % \node[anchor=north west, fill=blue!10] at (0.8, 0) {%
        \node[anchor=north west, fill=lightpink!40, minimum height=4.9cm, minimum width=0.99\columnwidth, rounded corners=4pt] at (0.2, 0) { };
    % Block 1
        \node[anchor=north west] at (0.75, -0.0) {\includegraphics[width=0.26\columnwidth]{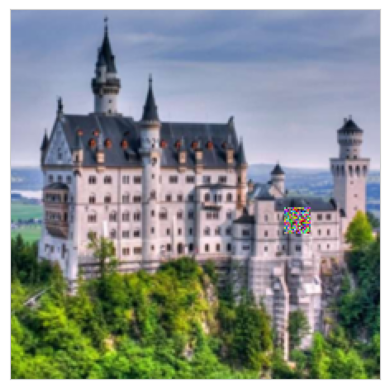} };   
    \node[rotate=90, anchor=center] at (0.5, -2.5) { \large \textsc{Known Triggers}};     \node[anchor=north west, text width=0.6\columnwidth, align=left] 
            at (0.39\columnwidth, -0.05) {%
                \begin{minipage}{0.99\linewidth}
                    \textbf{{\badnet~\cite{gu2017badnets}}} \hfill CA \hfill ASR \\ [0.5ex]
                    % \draw[thick] (0.45\columnwidth, -0.2) -- (0.9\columnwidth, -0.2);
                    \rule[\ht\strutbox]{\textwidth}{0.5pt}\\[-2ex]
                    \clip: \hspace{1.7cm} 57.5\% \hfill 99.2\%\\[0.2ex]
                    \roclip~\cite{yang2024robust}: \hspace{0.6cm} 47.4\% \hfill 75.1\%\\[0.2ex]
                    \cleanclip~\cite{bansal2023cleanclip}: \hfill 53.0\% \hfill 14.2\%\\[0.2ex]
                    \ours (ours):\hspace{1cm} 53.3\% \hfill \textbf{6.3\%} \\[0.2ex]
                \end{minipage}
        };
\draw[dashed] (1.0, -2.5) -- (0.99\columnwidth, -2.5);
    % Block 2
  \node[anchor=north west] at (0.75, -2.55) {\includegraphics[width=0.26\columnwidth]{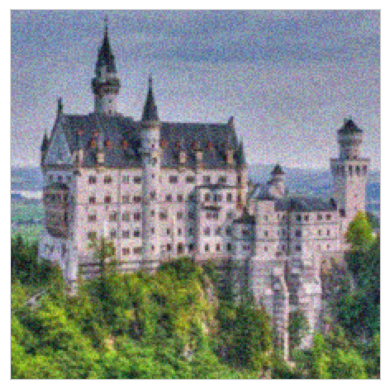} };   
    \node[anchor=north west, text width=0.6\columnwidth, align=left] 
            at (0.39\columnwidth, -2.55) {%
                \begin{minipage}{0.99\linewidth}
                    \textbf{{\blended~\cite{chen2017targeted}}} \hfill CA \hfill ASR \\[0.5ex]
                    \rule[\ht\strutbox]{\textwidth}{0.5pt}\\[-2ex]
                    \clip: \hspace{1.7cm} 57.7\% \hfill 99.4\%\\[0.2ex]
                    \roclip~\cite{yang2024robust}: \hspace{0.6cm} 47.9\% \hfill 1.5\%\\[0.2ex]
                    \cleanclip~\cite{bansal2023cleanclip}: \hfill 53.4\% \hfill 19.5\%\\[0.2ex]
                    \ours (ours):\hspace{1cm} 53.6\% \hfill  \textbf{0.0\%} \\[0.2ex]
                \end{minipage}
        };
        % \draw[thick] (0, -5.3) -- (0.95\columnwidth, -5.3);
    \node[anchor=north west, fill=lightblue!40, minimum height=4.9cm, minimum width=0.99\columnwidth, rounded corners=4pt] at (0.2, -5.0) { };
    % Block 1
        \node[anchor=north west] at (0.75, -5.0) {\includegraphics[width=0.26\columnwidth]{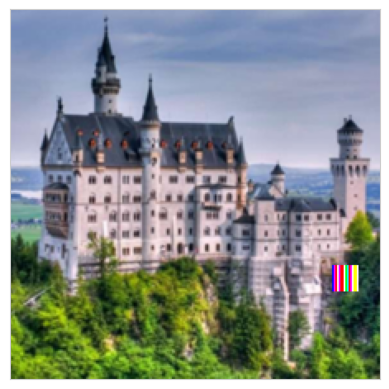} };   
    \node[rotate=90, anchor=center] at (0.5, -7.45) { \large \textsc{Proposed Triggers}};     \node[anchor=north west, text width=0.6\columnwidth, align=left] 
            at (0.39\columnwidth, -5.07) {%
                \begin{minipage}{0.99\linewidth}
                    \textbf{{\badnet-Stripes}} \hspace{0.5cm} CA \hfill ASR \\[0.5ex]
                    \rule[\ht\strutbox]{\textwidth}{0.5pt}\\[-2ex]
                    \clip: \hspace{1.7cm} 57.6\% \hfill 99.8\%\\[0.2ex]
                    \roclip~\cite{yang2024robust}: \hspace{0.6cm} 48.2\% \hfill 82.0\%\\[0.2ex]
                    \cleanclip~\cite{bansal2023cleanclip}: \hfill 53.0\% \hfill 62.3\%\\[0.2ex]
                    \ours (ours):\hspace{1cm} 53.0\% \hfill \textbf{42.4\%} \\[0.2ex]
                \end{minipage}
        };
\draw[dashed] (1.0, -7.5) -- (0.99\columnwidth, -7.5);
    % Block 2
  \node[anchor=north west] at (0.75, -7.55) {\includegraphics[width=0.26\columnwidth]{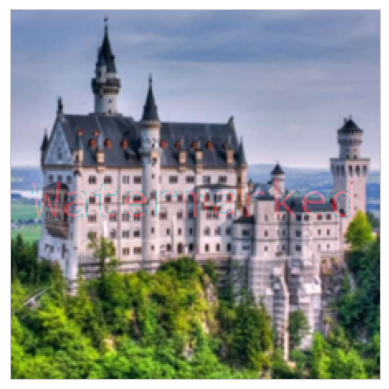} };   
    \node[anchor=north west, text width=0.6\columnwidth, align=left] 
            at (0.39\columnwidth, -7.55) {%
                \begin{minipage}{0.99\linewidth}
                    \textbf{{\blended-Text}}  \hspace{0.75cm} CA \hfill ASR \\[0.5ex]
                    \rule[\ht\strutbox]{\textwidth}{0.5pt}\\[-2ex]
                    \clip: \hspace{1.7cm} 56.9\% \hfill 95.6\%\\[0.2ex]
                    \roclip~\cite{yang2024robust}: \hspace{0.6cm} 47.2\% \hfill 59.1\%\\[0.2ex]
                    \cleanclip~\cite{bansal2023cleanclip}: \hfill 53.3\% \hfill 42.4\%\\[0.2ex]
                    \ours (ours):\hspace{1cm} 53.4\% \hfill \textbf{18.1\%} \\[0.2ex]
                \end{minipage}
        };
\end{tikzpicture}
\vspace{-0.2cm}
\caption{\textbf{PAR cleans better than previous methods.} We show clean accuracy (CA) and attack success rate (ASR) for the poisoned model (\clip) and the model after cleaning with \roclip~\cite{yang2024robust}, \cleanclip~\cite{bansal2023cleanclip} and our \ours. While \cleanclip and \roclip work well for known triggers, they perform worse for our novel (harder) structured triggers with \roclip suffering the most degradation in CA. PAR is the best backdoor defense across attacks and triggers that maintains highest CA.}
\label{fig:teaser}
\vspace{-1mm}

\end{figure}

% \fra{the first part of this paragraph, which i think should describe backdoors for CLIP, could read better}
Multi-modal models are trained on large (billions of samples) web-scraped datasets~\cite{schuhmann2022laion, gadre2024datacomp}. At such a large scale it is infeasible to curate samples, exposing them to serious security threats.
Backdoor attacks~\cite{gu2017badnets, chen2017targeted, liang2024badclip}, usually in the form of hidden triggers designed to hijack the model's behavior, pose one such threat. They inject malicious samples into the training/fine-tuning dataset to change the models output at test-time under the influence of such triggers. It is shown in~\cite{carlini2024poisoning} that poisoning web-scraped datasets via backdoors is practical in a very cheap manner.
Similarly, \clip models are very vulnerable to backdoor attacks, e.g. \cite{carlini2024poisoning, liang2024badclip} demonstrate that a poisoning rate as low as 0.01\% is sufficient for successful backdoor injection. Since training \clip models is very expensive in terms of dataset size and compute, cleaning it (getting rid of learnt backdoor) by training from scratch, \eg by detecting poisoned training samples, is not feasible. %Furthermore, cleaning from scratch requires access to millions of clean samples which is not guaranteed in the age of web-scraped training datasets. 
%Hence, our focus is specifically on cleaning poisoned \clip models via fine-tuning, with the help of a small (much less than the size of the training data) set of data.
\mh{Hence, our focus is %specifically 
on cleaning poisoned \clip models via fine-tuning, with the help of a small clean set of data, much smaller than the training data.}
%As obtaining even small sets of real, clean data might be costly, we further show that using only synthetic data is sufficient to clean poisoned \clip models. 

% \begin{figure}[t]
% \centering
% \includegraphics[width=0.95\linewidth]{figures/cleanclip_v_antiback_eta_smoothed.pdf}
% \caption{\textbf{ASR v Clean accuracy trade-off for cleaned RN50 with \badnet-Stripes.} We plot attack success rate (ASR) against clean accuracy on \imnet for different strength of the uni-modal augmentation loss of \cleanclip and different threshold ($\tau$) for our \ours loss. \cleanclip is not able to clean the model for the structured backdoor random stripes as stripes are quite different from the employed augmentation set. In contrast \ours can clean all tested backdoor attacks ranging from (structured) random to deterministic triggers in form of watermarks, triangles etc.}
% \label{fig:asr-ca-trade-off}
% \end{figure}

Several existing works
%Techniques using strong augmentations have been used extensively
rely on strong augmentations to remove backdoors from both uni-modal image classifiers~\cite{lineural, doan2023defending, borgnia2021strong} and multi-modal models~\cite{bansal2023cleanclip, yangbetter}: 
% \fra{better connect to previous sentence}
 %In the multi-modal setting,
 among these, \cleanclip~\cite{bansal2023cleanclip}, \roclip~\cite{yang2024robust} and \safeclip~\cite{yangbetter} aim at cleaning poisoned \clip models.
% This is justified as strong augmentations like AutoAugment (used by \cleanclip) may already encode similar perturbations as used in the backdoor trigger. 
Although for standard image classifiers using augmentations like horizontal flip, color based operations, cutout~\cite{devries2017improved}, etc., is not problematic as the class label is preserved, the same is questionable for CLIP as text captions encode more than just class information. For instance,
% \fra{i'd skip the numbering of the list, these are just example} \textit{(i)} 
a caption describing the relative position of two objects will be incorrect after horizontal flipping, %of the image
or %a caption describing the color of the image will also be incorrect after color altering operations
the colors mentioned in the caption will not match those in the image after color altering operations. Thus using strong augmentations for image-text paired data can be detrimental for standard performance of \clip.
% \fra{the jump from the discussion on augmentation to bypassing the existing methods is a bit unexpected... maybe this discussion is more relevant when introducing our method, to say that we design it in such a way to avoid the usage of heavy augmentations}\nds{updated a bit}

% Strong augmentations like AutoAugment~\cite{cubuk2019autoaugment} are fundamental parts of backdoor cleaning methods like % However, more importantly we show that known fine-tuning cleaning methods like
% \cleanclip~\cite{bansal2023cleanclip} and \roclip~\cite{yang2024robust}. 
% which are based on strong augmentations like AutoAugment~\cite{cubuk2019autoaugment} 
In this work, we first show that %cleaning methods 
\cleanclip~\cite{bansal2023cleanclip} and \roclip~\cite{yang2024robust}, cleaning methods using strong augmentations like AutoAugment~\cite{cubuk2019autoaugment}, can be bypassed by using simple structured triggers in known frameworks for backdoor attacks such as \badnet~\cite{gu2017badnets} and \blended~\cite{chen2017targeted}.
We hypothesize the reason for this is %the fact 
that structured triggers are unrelated to the image changes done by the augmentation operations.
Thus %these cleaning techniques yield a false sense of security as an effective backdoor defense should work for all possible backdoors as the backdoor is unknown to the defender.
these cleaning techniques yield limited security, as they implicitly assume some knowledge of the trigger used by the attacker, which is %instead 
unknown in practice.

Then, we propose a simple yet effective fine-tuning stage method for backdoor cleaning, termed \ours, short for \textit{Perturb and Recover}, which does not rely on strong augmentations. In particular, the training objective in \ours leverages a term to perturb the image and text encoders in \clip away from their original poisoned state: making the model ``forget'' the spurious correlations between the trigger and the target label it had learned during poisoning. Moreover, we add the standard \clip loss to preserve the initial clean performance. 
Overall, this approach allows us to achieve high cleaning performance across a variety of tasks across \clip models.
Finally, as obtaining even a small set of real, clean data might be costly, we show that using only synthetic data is sufficient to clean poisoned \clip models with \ours.

\noindent\textbf{Contributions.}
% \textbf{Contributions.} 
In summary, our main contributions are%as follows
:
\begin{itemize}
    \item we show that backdoor defenses for CLIP relying on data augmentations like \cleanclip~\cite{bansal2023cleanclip} and \roclip~\cite{yang2024robust} are easy to break %in black-box setup 
    with simple, structured %high-frequency
    triggers like stripes, triangles, overlayed text, see~\cref{fig:teaser}.
    \item we propose \ours, a simple but very effective fine-tuning scheme for cleaning \clip models from arbitrary backdoor triggers.
    %based on unlearning that cleans poisoned pre-trained CLIP models of backdoors.  
    Via extensive experiments across encoder architectures (ResNet50, \vit-B/L), models (CLIP, SigLip), backdoor attacks and downstream tasks (zero-shot classification, retrieval) we show the efficacy of the proposed technique.
    \item we show in the realistic setting where no real data is available, how synthetic (via text-to-image models) data can be leveraged to remove backdoors effectively.
    % \item In the plausible setting where sufficient clean data is lacking, we show how to leverage synthetic data with the proposed cleaning scheme effectively.
    % \item we show that employing CLIP models with backdoors in a vision-language model (VLM) like Llava 1.5 can yield significant performance drops in tasks like visual-question-answering (VQA) but cleaning with \ours is effective for the VLM as well. 
\end{itemize}

\section{Related Work}\label{sec:relatedwork}

A backdoor attack is done by poisoning a fraction (usually $ < 1\%$) of the training/fine-tuning datasets with specific triggers. A model trained on such a poisoned dataset changes in such a way that when an image containing the backdoor trigger is given as input the model acts in a targeted fashion, \eg in image classification it produces a certain target class (\eg, banana, refrigerator etc). Backdoor attacks have been studied \mh{in} a variety of setups \mh{to} highlight security vulnerabilities: image classification~\cite{chen2017targeted, gu2019badnets}, multi-modal models~\cite{liang2024badclip, carlini2021poisoning, yang2023data}, \mh{self-supervised learning}~\cite{jia2022badencoder, wang2024craft} and federated learning~\cite{bagdasaryan2020backdoor}.
On the positive side, backdoor triggers have also been used in applications like watermarking~\cite{li2020invisible, gu2022watermarking}, privacy protection~\cite{hintersdorf2024defending}, and copyright protection~\cite{wangstronger}. We focus on backdoor attacks as a security threat.

\textbf{Backdoor attacks.} In general, \textit{black-box} backdoor attacks assume no knowledge of the model to be poisoned but are unrestricted in the choice of the triggers. A large variety of attacks have been used, \eg random noise based patches like \badnet~\cite{gu2017badnets}, overlayed noise as in \blended~\cite{gu2017badnets}, containing natural phenomena like reflection~\cite{liu2020reflection}, sinusoidal signal with small frequency (SIG~\cite{barni2019new}). 
Another type are \textit{gray-box} backdoor attacks, which assume some access to the model for generating their trigger. WaNet~\cite{nguyen2021wanet} requires access to the model parameters in order to generate the warped trigger based on Fourier transform. For the multi-modal setup, \badclip~\cite{liang2024badclip} samples boundary images (using cosine similarities b/w clean images and target text) in order to optimize the trigger patch and poison while trying to keep the model parameters close to the original ones. 

% \subsection{Backdoor Cleaning methods}
\textbf{Uni-modal defenses.} A lot of different backdoor cleaning methods are available for image classifiers. A part of these focus on cleaning the model by leveraging clean datasets without adding any additional parameters~\cite{zhu2023enhancing, lineural, wu2021adversarial}  while some add additional parameters on top~\cite{doan2023defending, zhu2024neural}. In~\cite{li2021neural}, the backdoored model is fine-tuned on clean data for a few epochs and then used as a teacher to guide the cleaning process. In~\cite{zhu2024neural}, an additional layer is added in the middle/latter layers of the backdoored model before fine-tuning.~\cite{gao2023backdoor} use a dataset during cleaning which assumes access to poisoned data. Uni-modal cleaning methods can not directly be used for multi-modal models like \clip as they do one of the following, \textit{(i)} warrant access to some poisoned data which we believe is unrealistic for the defender, \textit{(ii)} introduce additional parameters which would change the embedding of the original \clip %unless large amounts of clean data is not available.
 model or \textit{(iii)} require large amounts of clean training data which is hard to obtain.

\textbf{Multi-modal defenses.}
Not many works have focused on cleaning poisoned multi-modal models. Recently, in \roclip~\cite{yang2024robust} clean training of a \clip model under poisoned training dataset was done with contrastive loss by using augmented images in conjunction with text retrieved from an adaptive pool.
% \fra{it's not clear what roclip does, even at high level}. 
SafeCLIP~\cite{yangbetter}, builds on \roclip and dynamically clusters the training set into clean and poisoned samples during cleaning. These two techniques train from scratch (random initialized model) which we think is unrealistic for large foundation models such as CLIP. In this work, we consider the setting that one has some clean data (real or synthetic) and fine-tunes the poisoned \clip model for cleaning, hence the total cost for fine-tuning is two to three orders of magnitude smaller than training from scratch. In this setting
\cleanclip~\cite{bansal2023cleanclip} proposes to clean a poisoned pre-trained \clip model using heavy data augmentation during fine-tuning.~\cite{kuang2024adversarial} assumes prior knowledge of the backdoor trigger while cleaning, which we consider a rather unrealistic scenario. Our goal is to work without any knowledge about the employed backdoor trigger. \mh{Recently, other fine-tuning \cite{xun2024cleanerclipfinegrainedcounterfactualsemantic,zhang2024defendingmultimodalbackdooredmodels} or detection methods \cite{niu2024bdetclipmultimodalpromptingcontrastive} have been proposed 
 but no code is available. Our code and models are available, but we don't cite them to retain anonymity.}

% \addi{Although in the main part of the paper we focus on cleaning poisoned pre-trained models, initial results for cleaning from scratch using \ours can be found in~\cref{app:safeclip}.}

\begin{figure*}[!t]
\centering
\includegraphics[width=0.98\linewidth]{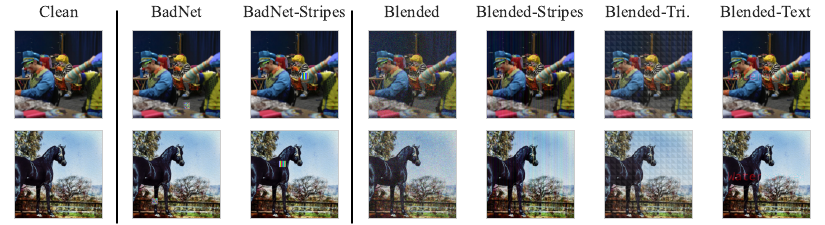}
\caption{\textbf{Visualizing different backdoor patterns.} Standard \badnet~\cite{gu2017badnets} and \blended~\cite{chen2017targeted} use Gaussian noise as a trigger, we replace the noise with random stripped pattern for \badnet termed \badnet-Stripes. For the \blended attack, we further replace the random noise with stripes, low contrast triangles (Blended-Tri.) and ``Watermarked'' text (\blended-Text), more visualizations in~\cref{fig:vis-backdoor-app}. \textit{Note: this is a very small subset of possible structured patterns, and we believe similar other patterns would be equally effective}.}
\label{fig:vis-backdoor}
\end{figure*}

\section{%Novel backdoor triggers are effective against established Backdoor Cleaning Techniques
Bypassing Augmentation-Based Backdoor Removal via Structured Triggers
}
\label{sec:effective-backdoors}
%\nds{add formal definition of a backdoor}

% \fra{attack and trigger are used at times interchangeably, probably it should be clarified}
We refer to the process of adding a backdoor trigger (patch, text, etc) to input images and changing the associated caption with the target label as a backdoor attack. Using such an attack, malicious users can poison some fraction of a dataset generating a \textit{poisoned dataset}. For poisoning (training the model with poisoned dataset) a pretrained \clip model with backdoor triggers, we operate with the strong premise that \textit{the attacker has no knowledge/control over the training process and models, that means the poisoning purely happens by contaminating the training data}. To reduce compute requirements, we avoid poisoning from scratch and instead poison pretrained \clip models.

\subsection{Attack constraints and goals}
%same as last sentence of previous pargraph% We focus on black-box backdoor attacks, where the attacker has no knowledge about the model but can only poison the the training data. In particular, f
For multi-modal models like \clip trained on massive amounts of web-scraped data, it is possible to manipulate training data as demonstrated by \cite{carlini2024poisoning} who got control of a fraction of LAION-400M~\cite{schuhmann2022laion} or Coyo-700M~\cite{kakaobrain2022coyo} at low cost. The goal of the attacker is to generate a poisoned dataset such that the model behaves normally on nominal images but yields a pre-defined behavior in presence of the backdoor trigger. Most backdoor triggers are not clearly visible or are hard to identify, as they correspond to normal variations of the image, \eg due to noise or watermarks in the image.
As an attacker is free to choose any backdoor trigger, it is important that mechanisms to remove backdoors work across different poisoning methods and triggers. In the following we introduce novel structured triggers for two common type of attacks. 

% Later, we show that recent defense techniques like \cleanclip are not effective against them \fra{where?}.

\subsection{Structured triggers for \badnet and \blended backdoor attacks}
\label{sec:proposed-triggers}

The \badnet~\cite{gu2017badnets} backdoor attack uses a localized trigger: a small patch of fixed random noise added at a random position in the image. In contrast, for standard \blended~\cite{chen2017targeted} attack, random noise ($N$) is overlayed on the original image ($I$) using the following convex sum (with $n_c=0.2$) to generate the backdoored images
        \begin{equation}
        (1-n_c) \times I + n_c \times N. \label{eqn:convex-sum}
    \end{equation}
We propose three novel types of triggers, visualized in~\cref{fig:vis-backdoor}, which can be integrated in either \badnet or \blended poisoning methods. We aim at replacing the random noise based patterns with more structured ones, which are unlikely to overlap with the augmentations used by backdoor removal methods.
In detail, we introduce

% In the following we replace the trigger pattern in th
% This section outlines the core design of our proposed backdoor triggers and how they can be used with known backdoor attacks. 
% The proposed triggers can be used in \badnet~\cite{gu2017badnets} style patch based attacks or overlayed on the full image similar to the \blended~\cite{chen2017targeted} attack. We show that recent defense techniques like \cleanclip are not effective against them.

% and they reduce the efficacy of the only fine-tuning based backdoor cleaning method ( \cleanclip~\cite{bansal2023cleanclip}) by several folds.

%\subsection{Structured triggers}

\begin{itemize}[parsep=3pt, topsep=0pt]
    \item \textbf{\badnet-Stripes.} We replace the standard Gaussian noise used in~\cite{gu2017badnets, bansal2023cleanclip} with stripes of width 1 pixel, whose color is randomly sampled from the corners of the color cube.
    % \fra{true?}.
    The size of the patch depends on the resolution.
    % The size of the patch is kept 16x16 \fra{this depends on the resolution i guess} as is common in the literature \cite{bansal2023cleanclip, liang2024badclip}.
    %for a 224$\times$224 image.
    
    \item \textbf{\blended-Stripes.} 
    We replace random noise based $N$ in~\blended with randomly sampled colored stripes of pixel-width 1 in~\cref{eqn:convex-sum}. The weight  $n_c$ is set to $0.03$, higher values would yield more visible triggers.
    
    \item \textbf{\blended-Triangles.} In the~\blended attack framework, we use low contrast isosceles triangles as $N$ instead of random noise. The triangles have a side of $14$ pixels, and $n_c$ is set to $0.15$ in~\cref{eqn:convex-sum}. This trigger is more visible but simulates a type of watermark. 
    
    \item \textbf{\blended-Text.} This backdoor is a classical watermark. We use red-colored text ``Watermarked'' using the NotMono font. The poisoned image is generated using~\cref{eqn:convex-sum} for all pixels covered by the text with $n_c = 0.5$ and keeps the original image $I$ elsewhere.
\end{itemize}
The proposed triggers are visualized in~\cref{fig:triggers} in \mh{the} Appendix. In the next section, we show how and why known cleaning methods like \cleanclip and \roclip can be bypassed by the proposed triggers and how the proposed cleaning method, \ours ameliorates the problem.

\section{Perturb and Recover: A Backdoor Defense without Bias to Specific Triggers}
\label{sec:par}
Most backdoor defenses techniques work by unlearning the backdoor trigger. We briefly review  the existing \cleanclip~\cite{bansal2023cleanclip} which has shown strong cleaning performance on existing backdoor attacks. However, we argue that \cleanclip is no longer effective when the backdoor trigger is uncorrelated with their cleaning procedure. This motivates our \ours which is a simple but effective cleaning technique that in contrast to \cleanclip (and similarly \roclip) has no bias towards a certain set of backdoor triggers. All backdoor defenses assume to have access to a set of clean data 
which is much smaller than the original data the \clip model has been trained on ($250k$ clean image/text pairs, this is $1/1600$ of the $400M$ the OpenAI \clip models \cite{radford2021learning} have been trained on). In~\cref{sec:synthetic-data} we show that even synthetic data is sufficient to clean from backdoors.

\subsection{Why \cleanclip is not effective against structured backdoor triggers}
% \subsection{\cleanclip and \roclip are not effective against structured triggers}
% \fra{move this to next subsection, as it's not introduction anymore} 
Both \cleanclip and \roclip use as part of the optimized loss, the contrastive clip loss term to retain good performance of the pretrained model which is given for  $(x_I^n,x_T^n)_{n=1}^B$ of images and text in a batch $B$ by:
\begin{align}
\label{eq:cliploss}
\L_{\text{CLIP}}=&\frac{-1}{2|B|}\sum_{n=1}^{|B|} \Bigg[ \log\left(\frac{\exp(\inner{\phi(x_I^n),\psi(x_T^n)})}{\sum_{k=1}^{|B|}\exp\left(\inner{\phi(x_I^n),\psi(x_T^k)}\right)}\right) \nonumber \\ & + %\frac{1}{2|B|} \sum_{n=1}^{|B|} 
\log\left(\frac{\exp(\inner{\psi(x_T^n),\phi(x_I^n)})}{\sum_{k=1}^{|B|}\exp\left(\inner{\psi(x_T^n),\phi(x_I^k)}\right)}\right) \Bigg], 
    \end{align}
where $\phi(.)$ and $\psi(.)$ are the normalized embeddings of the vision- and text-encoder of the CLIP model.

 % A recently proposed fine-tuning method that cleans poisoned \clip models is 
 
\cleanclip~\cite{bansal2023cleanclip} fine-tunes using a sum of $\L_{\text{CLIP}}$ loss and $\L_{\text{UniAug}}$ term, which is formed of two uni-modal self-supervised terms (image-image, and text-text) which are doing contrastive learning with augmented images and text:
\begin{align}
\label{eq:uniaugloss}  
\L_{\text{UniAug}}=& \frac{-1}{2|B|} \sum_{n=1}^{|B|}\Bigg[ \log\left(\frac{\exp(\inner{\phi(x_I^n),\phi(\textcolor{newpink}{\tilde{x}_I^n})})}{\sum_{k=1}^{|B|}\exp\left(\inner{\phi(x_I^n),\phi(\textcolor{newpink}{\tilde{x}_I^k})}\right)}\right) \nonumber \\ +& 
% \frac{1}{2|B|} \sum_{n=1}^{|B|}  
\log\left(\frac{\exp(\inner{\psi(x_T^n),\psi(\textcolor{newpink}{\tilde{x}_T^n})})}{\sum_{k=1}^{|B|}\exp\left(\inner{\psi(x_T^n),\psi(\textcolor{newpink}{\tilde{x}_T^k})}\right)}\right)\Bigg].
\end{align}
 %where $(x_I^n,x_T^n)$ are the clean image-text pairs, $\phi$ and $\psi$ are the vision- and text-encoder of the CLIP model.
Finally, \cleanclip uses as objective
\[ \L_{\text{\cleanclip}} = \L_{\text{CLIP}}+\lambda\, \L_{\text{UniAug}},\]
where they set $\lambda=1$. 
In Eq.~\eqref{eq:uniaugloss}, $\textcolor{newpink}{\tilde{x}_I}$ and $\textcolor{newpink}{\tilde{x}_T}$ denote the augmented versions of the original image $x_I$ and text $x_T$ respectively, which are generated by the strong AutoAugment~\cite{cubuk2019autoaugment} for images and EDA~\cite{wei2019eda} for text. The $\L_{\text{UniAug}}$ term in~\cref{eq:uniaugloss} allows the vision and image encoders to learn independently. By doing this, \cleanclip aims to break the spurious relations encoded during poisoning.

 \begin{figure}[t]
\centering
\includegraphics[width=0.95\linewidth]{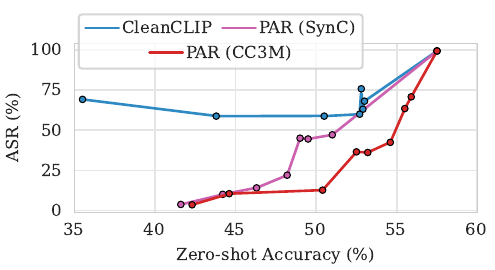}
\caption{\textbf{ASR v Clean accuracy trade-off for \badnet-Stripes cleaned RN50.} We plot attack success rate (ASR) against clean accuracy on \imnet for different strength of the uni-modal augmentation loss of \cleanclip and different threshold ($\tau$) for our \ours loss with clean (\ccm) and synthetic (SynC) data. \cleanclip is unable to clean the model for the proposed ``Stripes'' trigger pattern, which is quite different from the employed augmentation set. In contrast \ours completely cleans the model from the backdoor even with just synthetically generated (SynC) clean data.}
\label{fig:asr-ca-trade-off}
\vspace{-2mm}
\end{figure}

The same strong augmentations (AutoAugment, EDA) are also used for \roclip. Where for every $\K$ steps of standard contrastive training, one step of training is done using augmented versions of images and text retrieved from a dynamic retrieval pool. We note that AutoAugment is an ensemble of very strong augmentations and leads to heavy distortion of the image. While these heavy distortions seem to work well for the random noise patterns of \badnet and \blended (see \cleanclip and \roclip in~\cref{fig:teaser} and~\cref{tab:structured-patterns}), most likely as random noise is included in the set of augmentations, they are not effective against the structured backdoor trigger patterns we have introduced as they are still present after augmentation operations.
% \fra{i'd put this earlier, right after \eqref{eq:uniaugloss}, to finish to introduce \cleanclip, then discuss its weaknesses}

In~\cref{fig:asr-ca-trade-off}, we show Attack Success Rate (ASR) versus clean zero-shot performance of \cleanclip when increasing $\lambda$ for \badnet when using our structured trigger ``Stripes''. If $\L_{\text{UniAug}}$ was able to clean the backdoor, ASR should eventually decrease when increasing $\lambda$. However, even for very large $\lambda$ which affects clean zero-shot performance negatively, the ASR stays more or less constant. 

% \fra{not sure this should be here, or rather after you introduce \ours} In contrast our \ours (explained in next subsection) is able to clean the backdoor completely at the price of a small reduction in clean performance with both real data (\ccm) or completely synthetic data (SynC).

In practice, a backdoor cleaning technique should work for any possible backdoor attack and its employed trigger, as these are unknown. Thus while \cleanclip works reasonably well for some attacks, we do not consider it a reliable cleaning technique.  \mh{In \cref{app:cleaning-methods} we show that CleanCLIP can clean a \blended-stripes model if the augmentation ``stripes'' is added to their augmentation set, but this does not generalize to other structured triggers as CleanCLIP still fails on \blended-Text. In practice, the trigger is unknown and adding all potential structured triggers to the augmentation set is infeasible.}
Also \badclip~\cite{liang2024badclip} has shown that \cleanclip can be bypassed, but their trigger generation process requires access to the model to be poisoned. %, which is not always possible. 
 Having shown that \mh{cleaning based on} strong augmentations 
%lead to weak cleaning methods, 
\mh{is not reliable}, next we introduce our cleaning method which does not utilize any strong augmentations.  
%already talk about this in related work% Note, a trivial defense is to add the backdoor trigger as an augmentation during fine-tuning (as done in~\cite{kuang2024adversarial}) but knowing the trigger beforehand is not a realistic setup.
% \fra{i think this paragraph might be used to connect this section with the next, e.g. we have discussed that heavy augmentations have limitations, then we introduce a method which doesn't rely on them}

 % The proposed patterns have structured noise: high-frequency stripes, structured triangles and text. The spurious correlations the model has thus learned during poisoning might be highly different to the ones encoded using random noise. We believe this causes cleaning methods like \cleanclip to fail under such triggers, as it uses AutoAugment for augmentations and it does not comprise in high-frequency patterns as the proposed triggers. 
 
\begin{figure*}[t]
\centering
\begin{tikzpicture}

 \node[anchor=north west, minimum height=1.3cm, minimum width=0.94\linewidth, rounded corners=8pt] at (0.2, 0) { };
\node[anchor=north west] at (1.0, -0.0) {\includegraphics[width=0.94\linewidth]{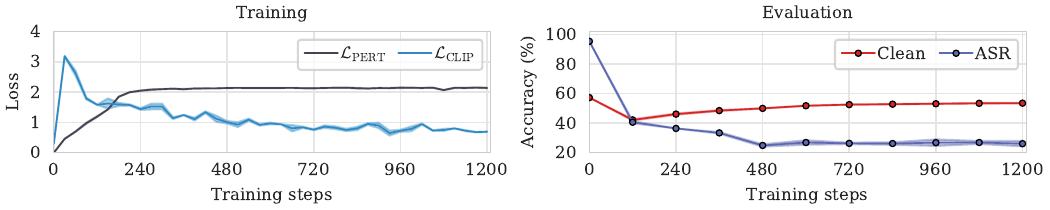} };   
\node[rotate=90, anchor=center] at (0.6, -1.6) { \textsc{\small Training \ours}};     
\node[anchor=north west, minimum height=4.0cm, minimum width=0.96\linewidth, rounded corners=8pt] at (0.1, -2) { };
\node[anchor=north west] at (1.2, -3.5) {\includegraphics[width=0.92\linewidth]{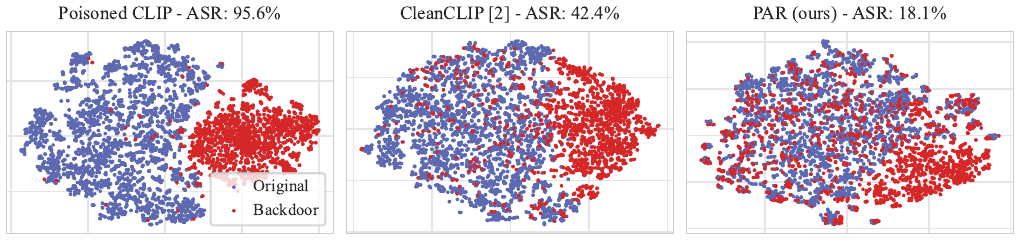} };   
\node[rotate=90, anchor=center] at (0.65, -5.75) {\textsc{\small t-SNE embeddings}}; 
\end{tikzpicture}
\vspace{-0.6cm}
\caption{\textbf{Training dynamics of \ours and visualizations of image embeddings across cleaning methods for \blended-Text poisoned RN50.} In the top left plot, we show how the $\L_{\text{CLIP}}$ and $\L_{\text{PERT}}$ ($\tau=2.15$) loss terms develop over training steps (evaluated every 25 steps) for \blended-Text poisoned RN50. Even though the schedule was optimized for \badnet-Stripes poisoned RN50, in the top right plot, we see how the training schedule generalizes by plotting clean accuracy and ASR (evaluated on $10k$ samples from \imnet). In the bottom row, we visualize the t-SNE~\cite{tsnevandermaaten08a} projections of the same \blended-Text poisoned \clip model (left), the one finetuned by \cleanclip (middle), and finetuned by \ours (right). Overall \ours yields the best mixing of clean and backdoored samples. Better mix means the model sees the clean and backdoored samples similarly, which also translates to low ASR. Similar visualizations for other attacks can be found in App.~\ref{app:visualize}.}
\label{fig:train-tsne}
\vspace{-1mm}
\end{figure*}

\subsection{\ours: Perturb and Recover}
\label{sec:anti-back}
Our proposed defense \ours, Perturb and Recover, is based on a controlled model perturbation which is independent from the potential backdoor attack or trigger. We actively enforce that the vision and text embedding of the poisoned model is perturbed significantly and stays away from the original poisoned model. At the same time we recover prediction performance by minimizing the standard \clip loss $\L_{\text{CLIP}}$ on the clean data. In this way, we unlearn the spurious correlation of the poisoned model while trying to preserve the performance of the original \clip model.
 
We denote by $\phi_P(\cdot), \psi_P(\cdot)$ the normalized embeddings of the vision and text encoders of the poisoned model.
As objective for enforcing the perturbation we use the $\ell_2$-distance between the features of the poisoned and the cleaned model, for both the image ($S_{\phi}$) and text ($S_{\psi}$) encoders on a batch $B$ of clean data $(x_I^n,x_T^n)_{n=1}^{|B|}$, i.e. we compute
\begin{align*}
S_{\phi} =& \frac{1}{|B|}\sum_{n=1}^{|B|} \left\lVert \phi(x^n_I) - \phi_P(x^n_I) \right\rVert^2_2, \\
S_{\psi} =& \frac{1}{|B|}\sum_{n=1}^{|B|} \left\lVert \psi(x^n_T) - \psi_P(x^n_T) \right\rVert^2_2.
\end{align*}
% \begin{align*}
% S_{\phi} =& \frac{1}{|B|}\sum_{n=1}^{|B|} \left\lVert \frac{\phi(x^n_I)}{\norm{\phi(x^n_I)}_2} - \frac{\phi_P(x^n_I)}{\norm{\phi_P(x^n_I)}_2} \right\rVert^2_2, \\
% S_{\psi} =& \frac{1}{|B|}\sum_{n=1}^{|B|} \left\lVert \frac{\psi(x^n_T)}{\norm{\psi(x^T_I)}_2} - \frac{\psi_P(x^n_T)}{\norm{\psi_P(x^T_I)}_2} \right\rVert^2_2.
% \end{align*}
As one can assume that the spurious backdoor correlation is destroyed when the model is perturbed enough, we threshold $S_\phi$ and $S_\psi$ in the objective at $\tau$ and average them to get the perturbation term.
\begin{equation}
\label{eqn:unlearn}
\mathcal{L}_{\text{PERT}} = \frac{1}{2} \left(\mathbb{I}[S_{\phi} \leq \tau] \cdot S_{\phi} + \mathbb{I}[S_{\psi} \leq \tau] \cdot S_{\psi}\right)
\end{equation}
where $\mathbb{I}$ is the indicator function. Then the complete loss of \ours takes the form
\begin{equation}
\label{eqn:loss}
\L_{\text{PAR}} = \L_{\text{CLIP}} -  \mathcal{L}_{\text{PERT}}.
% \begin{split}
% \label{eqn:lunlearn}
%      \L_{\text{{Unlearn}}} &= \frac{1}{N}\sum_{n=1}^{N} \Biggl(
%      \min\{\norm{\frac{\phi_{S}(x_I^n)}{\norm{\phi_{S}(x_I^n)}_2} - \frac{\phi_{T}(x_I^n)}{\norm{\phi_{T}(x_I^n)}_2}}_2^2, \rho \}\Biggl) \\ & +  \frac{1}{N}\sum_{n=1}^{N} \Biggl( \min\left\lbrace \norm{\frac{\psi_{S}(x_T^n)}{\norm{\psi_{S}(x_T^n)}_2} - \frac{\psi_{T}(x_T^n)}{\norm{\psi_{T}(x_T^n)}_2}}_2^2, \rho \right\rbrace\Biggl)
%      \end{split}
     \end{equation}
We note that $S_\phi$ and $S_\psi$ are bounded by four and thus the objective is well-defined for minimization. Moreover, one can equally write $S_\phi$ (and similarly $S_\psi$) as
\[ S_\phi = 2 - \frac{2}{|B|}\sum_{n=1}^{|B|} \cos\left(\phi(x^n_I),\phi_P(x^n_I)\right),\]
and thus one can also see the objective as jointly minimizing CLIP loss and cosine similarity to the poisoned model. 
%where $\L_{\text{CLIP}}$ is the standard \clip loss~\cite{radford2021learning}. 
%$\L_{\text{\text{PERT}}}$ is naturally bounded above by 4. 
% The choice of normalized, squared $\ell_2$-distance in $\L_{\text{PERT}}$ comes from the fact that it has commonly been used in distillation literature, see~\citep{jiao2023learning, sariyildiz2024unic} and also for making CLIP models adversarially robust~\cite{schlarmann2024robustclip}. Other distance based losses like $\ell_1$, smooth-$\ell_1$~\cite{sariyildiz2024unic} would also work although for a different $\tau$. 
Once $S_\phi$ and $S_\psi$ are both larger than $\tau$, that is the embedding $(\phi,\psi)$ is sufficiently far away from that of the poisoned model $(\phi_P,\psi_P)$, one effectively only minimizes the CLIP loss $\L_{\text{CLIP}}$ and thus recovers the performance of the CLIP model.
During cleaning (training to get rid of the backdoor induced spurious correlations), we use two simple augmentations in Gaussian noise and CutOut with a very small patch. These augmentations do not significantly distort the main objects in the image,  details in ~\cref{app:exp-details}. The choice of these augmentation is justified in~\cref{tab:reb-aug}.
%do not alter the image content in any way (the  patches are very small and never hide main-object centric information in the image), details in ~\cref{app:exp-details}. 

% \fra{maybe this could be a separate subsection, as it's more experiments than method definition}
\subsection{Analysing \ours}

\textbf{Accuracy vs backdoor removal trade-off.}
In the cleaning process with \ours one can control the trade-off between keeping clean performance for non-backdoored images and having low attack success rate (ASR) using $\tau$. We have seen in~\cref{fig:asr-ca-trade-off} that \cleanclip is not able to effectively clean the model for \badnet with our trigger random stripes (BadNet-Stripes). Nevertheless, in order to allow a fair comparison we choose the threshold $\tau$ in~\cref{eqn:unlearn} so that we reach roughly the same zero-shot performance for our cleaned model with \badnet-Stripes as \cleanclip with their optimized parameters (which aim at retaining high clean performance). This value of $\tau=2.15$ is then used for all tested backdoor attacks with different triggers as well as for the other \clip encoders with \vit-B/32 architecture. As can be seen in~\cref{fig:asr-ca-trade-off}, choosing a larger $\tau$ for \ours can further reduce ASR at the price of a small loss in zero-shot accuracy. Unlike \cleanclip, \ours can completely clean the \badnet-Stripes trigger backdoored \clip model whether it is with clean data comprising of real (\ccm) or synthetic (SynC) images (see Sec.~\ref{sec:synthetic-data} for more on synthetic data).

\textbf{Training dynamics.}
In~\cref{fig:train-tsne}, we illustrate the training dynamics of the cleaning process of \ours for the model poisoned with the proposed \blended-Text backdoor. In the top left plot, we see that $\L_\text{CLIP}$ grows quickly at the start of training due to the perturbation term $\L_\text{PERT}$ but decays again once the perturbation loss $\L_\text{PERT}$ saturates at the threshold (black curve in the top left plot). At the same time increasing model perturbation leads to a decay of the ASR which decays further after the threshold is activated in the perturbation loss. We note that while it looks like as if $\L_\text{PERT}$ is constant after some time, the loss is activated from time to time for some batches pushing the model again away once it tends in the direction of the poisoned model. Even though the training schedule (detailed in~\cref{sec:experiments}) and $\tau$ were optimized for a model poisoned with \badnet-Stripes,  both generalize well to other (structured) triggers.

In the bottom row of~\cref{fig:train-tsne}
 we visualize for the \blended-Text poisoned and cleaned model the t-SNE~\cite{tsnevandermaaten08a} embeddings of the original and backdoored images. A well-separated embedding indicates that the model behaves differently for original and backdoored inputs whereas a mixed embedding indicates that the backdoor is removed. For the poisoned model, the embeddings form two disjoint clusters which are mixed up a bit after cleaning with \cleanclip. Whereas, in the rightmost plot 
a more homogeneous mixing of embeddings is achieved by \ours, clearly indicating its effectiveness in backdoor removal via the low ASR.

\section{Experiments and Discussion}
\label{sec:experiments}
\subsection{Experimental Details}
\label{sec:experimental-details}
\textbf{Poisoning.} As already stated, we assume no access/control over poisoning. Hence, irrespective of the backdoor attack, we always poison a model in the same fashion. We poison the OpenAI pretrained \clip models~\cite{radford2021learning} using different vision encoders (RN50, ViT-B/32) with a subset of \ccm~\cite{sharma2018conceptual} dataset. The default poisoning rate is set to 0.5\% ($2k$ poisoned / $400k$ clean samples), and poisoning is done for 5 epochs. Unless specified otherwise, the target label is set to ``banana'', which is prevalent in literature~\cite{bansal2023cleanclip, liang2024badclip}. More
% For generating triggers with existing attacks (\badnet~\cite{gu2017badnets}, \blended~\cite{chen2017targeted}, WaNet~\cite{nguyen2021wanet}), we follow the setup from~\cite{bansal2023cleanclip}.
details can be found in~\cref{app:exp-details}\mh{, further results on ViT-L/14 and SigLip~\cite{zhai2023sigmoid}  are in~\cref{app:additional-exp}.}%  and  in \cref{tab:reb-siglip}.}.

\begin{table*}[t]
\centering
\small
\tabcolsep=0.7pt
\extrarowheight=1.5pt
\newl=9mm
\newlc=8.5mm
\begin{tabular}{C{5.5mm} L{20mm} |*{8}{|C{\newlc} C{\newlc}}}
\toprule
& & \multicolumn{4}{c|}{\badnet} & \multicolumn{8}{c|}{\blended} 
% & \multicolumn{2}{c|}{\multirow{2}{*}{SIG~\cite{barni2019new}}} 
& \multicolumn{2}{c|}{\multirow{2}{*}{\makecell{\wanet\\\cite{nguyen2021wanet}}}} & \multicolumn{2}{c}{\multirow{2}{*}{\makecell{\badclip\\\cite{liang2024badclip}}}}
\\
 & & \multicolumn{2}{c|}{Rand.~\cite{gu2017badnets}} & \multicolumn{2}{c|}{Stripes} & \multicolumn{2}{c|}{Rand.~\cite{chen2017targeted}} & \multicolumn{2}{c|}{Stripes} & \multicolumn{2}{c|}{Triangles} & \multicolumn{2}{c|}{Text} & & & & \\
\cmidrule{2-18}

& Method & clean & ASR  & clean & ASR  & clean & ASR  & clean & ASR  & clean & ASR  & clean & ASR  & clean & ASR  & clean & ASR  \\ % & clean & ASR  \\
\addlinespace[0.5mm]
\toprule 
\addlinespace[0.5mm]
\rowcolor{NewGray} \cellcolor{white} \multirow{4}{*}[-0.4em]{\rotatebox{90}{\textcolor{cvprblue}{\imnet}}} & \clip & 57.5 & 99.2 & 57.6 & 99.8 & 57.7 & 99.4 & 57.6 & 95.6 & 57.4 & 85.7 & 56.9 & 95.6 
% & 56.6 & 60.4 
& 57.7 & 99.2 & 58.6 & 98.8\\
\addlinespace[0.5mm]
\cdashline{2-18}
\addlinespace[0.5mm]
&  \roclip~\cite{yang2024robust}  & 47.4 & 75.1 & 48.2 & 82.0 & 47.9 & 1.5 & 47.9 & 7.0 & 47.2 & 37.1 & 47.2 & 59.1 
% & 47.6 & 5.3
& 47.2 & 2.0 & -- & -- \\
& \cleanclip~\cite{bansal2023cleanclip}& 53.0 & 14.5 & 53.0 & 62.3 & 53.4 & 19.5 & 53.1 & 61.8 & 52.9 & 48.7 & 53.3 & 42.4 
% & 53.1 & 7.0 
& 52.9 & \textbf{0.0} & 53.8 & 40.1 \\
& \ours (ours) & 53.3 & \textbf{6.3} & 53.0 & \textbf{42.4} & 53.6 & \textbf{0.0} & 53.5 & \textbf{0.1} & 52.9 & \textbf{10.3} & 53.4 & \textbf{18.1} 
% & \textbf{53.5} & \textbf{0.1} 
& 54.4 & \textbf{0.0} & 53.4 & \textbf{30.4} \\
\addlinespace[0.5mm]
\midrule
\addlinespace[0.5mm]
\rowcolor{NewGray} \cellcolor{white} \multirow{4}{*}[-0.2em]{\rotatebox{90}{\textcolor{cvprblue}{COCO}}} & \clip & 73.3 & 99.4 & 73.1 & 99.9 & 73.2 & 99.3 & 73.2 & 98.2 & 73.1 & 98.2 & 73.2 & 97.4 
% & 72.5 & 35.4 
& 72.5 & 99.8 & 73.9 & 99.8\\
\addlinespace[0.5mm]
\cdashline{2-18}
\addlinespace[0.5mm]
& \roclip~\cite{yang2024robust} & 66.4 & 51.4 & 66.4 & 75.9 & 66.9 & 6.7 & 66.6 & \textbf{4.9} & 66.3 & 84.6 & 66.2 & 49.1 
% & 66.6 & 13.9 
& 66.2 & 13.0 & -- & -- \\
& \cleanclip~\cite{bansal2023cleanclip} & 69.9 & 19.4 & 69.7 & 52.1 & 70.4 & 44.3 & 69.8 & 76.0 & 69.8 & 71.6 & 70.2 & 32.5 
% & \textbf{70.1} & 11.8 
& 70.1 & 5.1 & 70.3 & 62.6 \\
& \ours (ours) & 70.5 &\textbf{6.5} & 70.2 & \textbf{20.9} & 70.4 & \textbf{1.6} & 70.7 & 5.0 & 70.3 & \textbf{24.1} & 71.3 & \textbf{11.4} 
& 71.1 & \textbf{2.3} & 70.7 & \textbf{31.2} \\
\bottomrule
\end{tabular}
\caption{\textbf{Comparing \ours with \cleanclip and \roclip}. For RN50 based \clip, we report clean and ASR (lower is better) performance for \imnet zero-shot classification and COCO text retrieval. Attack-wise lowest ASR is in \textbf{bold} and the poisoned \clip is \colorbox{NewGray}{highlighted}.}
\label{tab:rn50-novel}
\vspace{-1mm}
\end{table*}

\textbf{Cleaning.} Unless specified otherwise, a set of $250k$ samples (similar to~\cite{liang2024badclip, bansal2023cleanclip}) disjoint of the poisoning set from the \ccm dataset is used for cleaning. Details on hyper-parameters for \cleanclip and how \roclip was adapted to fine-tuning setup can be found in~\cref{app:exp-details}. For  \ours, we use the setup defined in~\cref{sec:empirical-loss} and clean for 10 epochs, other details
% on augmentations, and other training parameters 
can be found in App.~\ref{app:cleaning-methods}. \addi{In~\cref{app:extra-par}, we show that \ours also works effectively if small amount of backdoor data is still present in the $250k$ clean sample set.}

\textbf{Metrics.} We evaluate all \clip models for zero-shot classification accuracy on the test set of \imnet at the standard $224 \times 224$ resolution. We report both clean accuracy and the attack success rate - ASR (percentage of samples with the backdoor trigger for which the model outputs the target ``banana''). A task native to multi-modal models is retrieval. Hence, we also evaluate retrieval on $5k$ points from the validation set of COCO dataset, where we report clean top-5 text retrieval score and ASR - fraction of samples with the backdoor trigger for which the target is retrieved in any one of the top-5 retrieved captions.

\subsection{Empirical considerations for the \ours loss}
\label{sec:empirical-loss}
We choose $\tau$, the only parameter in $\L_{\text{PAR}}$ by doing a sweep over it for \badnet-Stripes poisoned ResNet50 and use the same value for all other attacks/encoders. As we want to go away from init point (in weight space), we devise a custom schedule that helps us achieve this efficiently. We start with a high LR of $3e-5$ that decays to a value of $3e-6$ linearly over half of total cleaning epochs. For the remaining half, it goes down to $1e-9$ with a cosine schedule. However \ours also works effectively for the standard cosine decaying schedule as used in \cleanclip, see~\cref{app:additional-exp}.

\subsection{Evaluating \ours against diverse attacks}
\begin{figure}[t]
\centering
\includegraphics[width=0.95\linewidth]{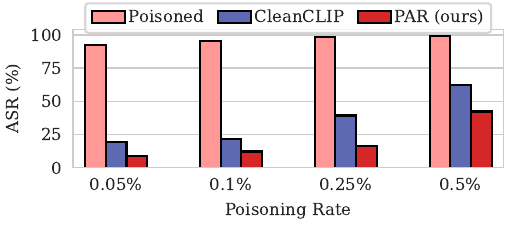}
\vspace{-1.8mm}
\caption{\textbf{ASR for different poisoning rates of \cleanclip and \ours for RN50.} Even at a lower poisoning rate of 0.05\%, \badnet-Stripes achieves 92\% attack success rate (ASR). Overall across all poisoning rates, \ours cleans better than \cleanclip.}
\label{fig:poison-rate}
\vspace{-5mm}
\end{figure}

As \clip with ResNet50 encoder has been used as standard test-bed for backdoor attacks and defenses, we first evaluate this. We test \ours, \cleanclip and \roclip against different backdoor attacks in~\cref{tab:rn50-novel}. 
\ours attains for all but one (\badclip) attack the best clean zero-shot ImageNet accuracy, whereas \roclip yields the worst.
Importantly, \ours outperforms in all cases both \cleanclip and \roclip in terms of ASR, and the difference between \ours and \cleanclip is as high as 61\% (\blended-Stripes).
Even though \roclip cleans better than \cleanclip, its clean performance is inferior. Similarly, \ours cleans better than \cleanclip also for lower poisoning rates, see~\cref{fig:poison-rate}.
For the recent ~\badclip attack, which assumes access to the model to be poisoned, \ours is very effective (since we use the poisoned model from~\cite{liang2024badclip} training \roclip in this setup was not possible). 
The same trend holds for top-5 retrieval for COCO, where \ours outperforms \cleanclip by as much as 65\% on ASR while maintaining similar clean performance. \roclip again yields the worst clean numbers. In~\cref{app:safeclip}, we discuss a comparison to SafeCLIP~\cite{yangbetter}.

\textbf{Adaptive attacks.}
\addi{As an adaptive attack against \ours, we re-poison the model cleaned with \ours. The idea is that then the original backdoored model is a potential solution for \ours when cleaning the re-poisoned model.~\cref{tab:repoison} in the Appendix shows that for Blended-Stripes \ours cleans the re-poisoned model perfectly and \ours does not ``fall back'' to the original backdoored model.}
%. The original backdoored model seems not to be an attractor in the optimization landscape.}
%
%this does not work.}

\begin{table*}[t]
\centering
\small
\tabcolsep=0.7pt
\extrarowheight=0.8pt
\newl=9mm
\newlc=9.4mm
\begin{tabular}{C{7mm} L{22mm} |*{7}{|C{\newlc} C{\newlc}}}

\toprule
& & \multicolumn{4}{c|}{\badnet} & \multicolumn{8}{c|}{\blended} 
% & \multicolumn{2}{c|}{\multirow{2}{*}{SIG~\cite{barni2019new}}} 
& \multicolumn{2}{c}{\multirow{2}{*}{\makecell{\badclip\\\cite{liang2024badclip}}}}
\\
 & &  \multicolumn{2}{c|}{Rand.~\cite{gu2017badnets}} & \multicolumn{2}{c|}{Stripes}  & \multicolumn{2}{c|}{Rand.~\cite{chen2017targeted}}& \multicolumn{2}{c|}{Stripes} & \multicolumn{2}{c|}{Triangles} & \multicolumn{2}{c|}{Text} & &  \\
\cmidrule{2-16}
& Method & clean & ASR & clean & ASR & clean & ASR  & clean & ASR  & clean & ASR  & clean & ASR  & clean & ASR  \\ % & clean & ASR  \\
\addlinespace[0.5mm]
\toprule 
\addlinespace[0.5mm]
\rowcolor{NewGray} \cellcolor{white} \multirow{3}{*}{\rotatebox{90}{\textcolor{cvprblue}{\imnet}}} & \clip & 59.6& 89.9 & 59.8 & 99.1 & 58.6 & 50.9 & 59.7 & 99.8 & 58.9 & 99.7 & 59.3 & 99.8  & 59.9 & 95.0\\
\addlinespace[0.5mm]
\cdashline{2-16}
\addlinespace[0.5mm]
& \cleanclip~\cite{bansal2023cleanclip} & 55.1 & 6.4 & 54.7 & 86.8  & 55.0 & 0.1 & 55.0 & 15.2 & 54.6 & 91.4 & 54.7 & 62.9 &  55.2 & 19.6 \\
& \ours (ours) & 55.5 & \textbf{0.1} & 54.4 & \textbf{50.1} & 55.5 & \textbf{0.0} & 54.6 & \textbf{0.1} & 54.9 & \textbf{15.9} & 56.1 & \textbf{37.3}  & 54.7 & \textbf{18.2}\\
\addlinespace[0.5mm]
\midrule
\addlinespace[0.5mm]
\rowcolor{NewGray} \cellcolor{white} \multirow{3}{*}{\rotatebox{90}{\textcolor{cvprblue}{COCO}}} & \clip & 72.2 & 85.6 & 72.5 & 98.9  &71.3 & 59.9& 72.6 & 99.8 &  71.9 & 99.9 & 71.8 & 99.8  & 72.9 & 96.8\\
\addlinespace[0.5mm]
\cdashline{2-16}
\addlinespace[0.5mm]
& \cleanclip~\cite{bansal2023cleanclip} & 70.3 & 11.2 & 70.5 & 84.7 & 70.4 & 2.9 & 70.6 & 22.4 &  70.2 & 98.5 & 70.6 & 55.4  & 70.1 & 28.5\\
& \ours (ours) & 69.4 & \textbf{1.4} & 69.5 & \textbf{51.6} & 69.4 & \textbf{1.5} & 69.4 & \textbf{1.7} &  70.0 & \textbf{91.5} & 70.2 & \textbf{45.4} & 70.6 & \textbf{27.1}\\
\bottomrule
\end{tabular}
\caption{\textbf{Comparing \ours with \cleanclip for \vit-B/32}.For \vit-B/32-\clip, we report clean and ASR (lower is better) performance for \imnet zero-shot classification and COCO text retrieval. Attack-wise lowest ASR is in \textbf{bold} and the poisoned \clip is \colorbox{NewGray}{highlighted}.}
\label{tab:vitb-novel}
\vspace{-3mm}
\end{table*}

\subsection{Changing the encoder}
From~\cref{tab:rn50-novel}, we omit the weakest attack in WaNet and evaluate the remaining ones for the \vit-B/32 based \clip model. Training setup is the same as for RN50. We omit \roclip as it had the worst clean-ASR trade-off for RN50 and it is a method native to cleaning models trained from scratch. From~\cref{tab:vitb-novel} for ViT-B/32, \ours has again a better backdoor removal rate than \cleanclip, improving by up to 37\% for ASR on \imnet while having similar clean performance. Even for COCO, \ours outperforms \cleanclip comprehensively in terms of ASR while always having nearly similar clean performance. Thus \ours generalizes across different architectures. We emphasize that one can reduce the ASR significantly for \ours by \mh{increasing} $\tau$, as can be seen in~\cref{tab:tau-vit}. Moreover, in~\cref{app:additional-exp}, we show \ours also works for a vision-language model trained with SigLip~\cite{zhai2023sigmoid} and larger models like \vit-L/14. 

% \textbf{Scaling \ours to \vit-L.} In~\cref{tab:vitl-exp}, we test if \ours scales out of the box (using same parameters as RN50) to larger models like \vit-L/14 based \clip. We poison and clean at resolution of $336\times336$ with a higher poisoning rate of 1\% and do cleaning only for 5 epochs, more details in~\cref{app:exp-details}. Even in this harder setting, \ours cleans for both \badnet-Stripes and \blended-Text better than \cleanclip while attaining similar clean performance. The clean accuracy as attained by \ours can be increased by adjusting $\tau$, see~\cref{tab:tau-vit}.

\subsection{Effective utilization of synthetic data}
\label{sec:synthetic-data}
Until now, we assumed access to 250k clean samples of image/caption pairs in \ccm.
However, getting 250k clean samples would come at a significant cost as one needs to  either \textit{(i)} detect the backdoored data, a difficult task on its own, or \textit{(ii)} label the data.
Thus we explore if synthetic data is a viable solution to simplify the backdoor removal.
%potential solution to such a problem is clean synthetic data generation via text-to-image models~\cite{rombach2022high}. 

In~\cite{hammoud2024synthclip}, a synthetic (image, caption) dataset, termed~\synth, is generated via text-to-image diffusion models~\cite{rombach2022high} and used to train \clip models at the price of slightly worse performance. We use a subset of 250k samples of \synth to clean poisoned models with \ours. 
%Firstly, similar to the real data (\ccm), \ours with \synth (SynC) dataset can completely clean a poisoned model with BadNet-Stripes, see~\cref{fig:teaser} yielding a trade-off curve similar to the real data with slightly decreasing clean performance.
Firstly, \cref{fig:asr-ca-trade-off} shows that \ours with \synth (SynC) can completely clean a poisoned model with BadNet-Stripes, and yields a trade-off curve similar to using real data (\ccm), with slightly reduced clean performance.
Moreover, in~\cref{tab:synthetic} we detail the backdoor removal rate of \ours with synthetic data.
We use the same threshold and training parameters as before with 250k/500k synthetic samples. For both \badnet-Stripes and \blended-Text, \ours brings \imnet ASR down significantly. The clean performance on \imnet is also slightly degraded, probably due to the drastic data distribution shift from the pre-training on real data coupled with the learning schedule. Interestingly, we obtain no reduction or even an improvement in clean performance at much smaller ASR for retrieval on COCO, potentially due to better captions in~\synth. Given the potential high cost of clean data, \mh{our} results show that synthetic data is a cheap and effective replacement \mh{with strong performance of \ours.}
%\ours shows strong backdoor removal abilities. 

%The degradation in clean performance on \imnet can be recovered to some extent by adding a fraction of real data to the synthetic one (this leads to a different point on the trade-off curve), where as low as $50k$ real data points seem to be enough. An additional benefit of using \synth data with good captions is the increase in clean retrieval performance on COCO. For instance, for \badnet-Stripes poisoned model using SynC with \ours yields a clean score of 71.0 which is better than the 69.5 one gets with real data while having 30\% better ASR. 

\section{Conclusion}
We show that know cleaning methods like~\cleanclip, \roclip can be bypassed by using our novel structured trigger patterns, due to their over-reliance on strong data augmentations. To overcome this we introduce \ours, a simple, intuitive and effective backdoor removal technique based on fine-tuning.~\ours cleans models from backdoors for all tested attacks across architectures/models effectively in comparison to the baselines. Finally, we show that synthetic data alone can effectively remove backdoor correlations in \clip models, reducing the need for costly real data.

\begin{table}[t]
\centering
\small
\tabcolsep=1.2pt
\newl=8mm
\newlc=10mm
\begin{tabular}{L{10mm} C{15mm} C{15mm} *{2}{|C{\newl} C{\newlc}}}
\toprule
& & & \multicolumn{2}{c}{\imnet} & \multicolumn{2}{c}{MS-COCO} \\
\cline{4-7} 
Method & Data & Samples & clean & ASR & clean & ASR  \\
\toprule
\addlinespace[0.5mm]
\multicolumn{6}{l}{\textbf{Attack:} \badnet-Stripes}\\
\midrule
% \rowcolor{NewGray}\clip & \ccm & 250k & 59.8 & 99.1 & 72.5 & 98.9\\
\ours & \ccm & 250k & 54.4 & 50.1 & 69.5 & 53.5 \\
\ours & \synth & 250k & 50.0 & 15.9 & 71.0 & 22.6\\
 \ours & \synth & 500k & 48.0 & 3.7 & 69.4 & 8.8 \\
\addlinespace[0.5mm]
\toprule
\addlinespace[0.5mm]
\multicolumn{6}{l}{\textbf{Attack:} \blended-Text}\\
\midrule
% \rowcolor{NewGray}\clip & \ccm & 250k & 59.3 & 99.8 &  71.8 & 99.8 \\
\ours & \ccm & 250k & 56.1 & 37.3  & 70.2 & 45.4 \\
\ours & \synth & 250k & 49.5 & 5.9 & 70.2 & 15.1 \\
\ours & \synth & 500k & 49.3 & 5.2 & 70.2 & 11.0 \\
\bottomrule
\end{tabular}
\caption{\textbf{Utilizing synthetic data for cleaning.} We clean \vit-B/32 poisoned \clip models using \ccm (CC) and \synth (SynC). \ours seems to effectively clean even with sythetic data.} 
\vspace{-1mm}
\label{tab:synthetic}
\end{table}

\clearpage 

{
    \small
    \bibliographystyle{ieeenat_fullname}
    \bibliography{main}
}

% WARNING: do not forget to delete the supplementary pages from your submission 
\clearpage
\setcounter{page}{1}
\appendix

\section*{Contents}
\begin{enumerate}
\itemindent=10pt
\item~\cref{app:exp-details} \ldots Experimental details and discussions
\item~\cref{app:additional-exp} \ldots Additional experiments 
\item~\cref{app:visualize} \ldots  More visualizations
   % \item 
\end{enumerate}

\section{Experimental Details and Discussions}
\label{app:exp-details}

In this section we detail the setup related to all the experiments conducted in this work.
We detail how we select training hyperparameters like batch size ($BS$), learning rate ($LR$), datasets used, optimizer, etc., for poisoning and cleaning across methods and models. All experiments were conducted on 4-8 A100 GPUs.

\subsection{Dataset}
\label{sec:app-poisoning-details}

\textbf{Poisoning.} For poisoning and cleaning we always use the \ccm dataset. For poisoning (adding backdoor to the model), we use $400k$ samples out of which $2k$ are poisoned. For experiments involving lower or higher poison rate only the number of poisoned samples was changed accordingly. The target caption associated with the trigger appends the target label to a template for \eg ``an image of \{target-label\}'' with the caption template randomly sampled from 80 zero-shot templates from~\cite{ilharco_gabriel_2021_5143773}. 

\noindent\textbf{Cleaning.} We use $250k$ samples also from \ccm, disjoint of the poisoning set. For experiments with synthetic data, we use the respective sized subsets from \synth dataset~\cite{hammoud2024synthclip}. For evaluation, we use the full validation set of \imnet~\cite{deng2009imagenet} and $5k$ samples from the COCO2014~\cite{lin2014microsoft} validation split. For poisoning, cleaning and evaluation of RN50 and \vit-B/32 based \clip models we use the $224 \times 224$ resolution, whereas for \vit-L/14 we employ a higher resolution of $336\times336$.

\subsection{Poisoning}
Irrespective of the model/attack, the poisoning is always carried out with a fixed cosine-decay schedule with peak $\text{LR}=1e-5$ trained for 5 epochs with $\text{BS}=256$. We employ the AdamW optimizer with default \texttt{PyTorch} hyperparameters and a weight decay of $1e-4$ and optimize with the standard \clip loss ($\L_{\text{CLIP}}$). All poisoning for \vit-B/32 and Resnet50 is done at the resolution of $224 \times 224$. For \badnet-Stripes we use a patch of size $16\times16$ as is common in literature~\cite{bansal2023cleanclip, liang2024badclip}. For other proposed triggers, we follow the setup listed in~\cref{sec:proposed-triggers}.

For known attacks (\badnet~\cite{gu2017badnets}, \blended~\cite{chen2017targeted}, SIG~\cite{barni2019new} and WaNet~\cite{nguyen2021wanet}), we follow the attack specific parameters from~\cite{bansal2023cleanclip}. In \badclip the patch ($16 \times 16$) is optimized on the model and the poisoning is also done with a specific training setup - this is contrary to our setup where we assume no knowledge/control over poison training process. For RN50, we use the patch and model from the their Github repository.\footnote{\href{https://github.com/LiangSiyuan21/BadCLIP}{https://github.com/LiangSiyuan21/BadCLIP}} As \badclip authors do not show results for ViT based \clip in their work, we optimize the trigger patch using their code with $5k$ auxiliary samples from \ccm for $50$ epochs with a batch size of $128$ and poison with our schedule for $5$ epochs. 

For poisoning \vit-L based \clip model, we use a 1\% poisoning rate and use a higher resolution of $336 \times 336$ as downstream use of \vit-L/14 \clip for \eg in LVLMs (LLaVa, CogVLM) is often done at this higher resolution. For \badnet-Stripes we use a patch of size $32\times32$. For \blended-Text, we scale the text ``Watermarked'' to span across the width of the image and use $n_c = 0.5$ in~\cref{eqn:convex-sum}. We keep the LR and schedule same as before and poison for 8 epochs. We found poisoning with 5 epochs was not as effective as for RN50 or \vit-B/32 based \clip.
\begin{figure}[!t]
\centering
    \includegraphics[width=0.98\linewidth]{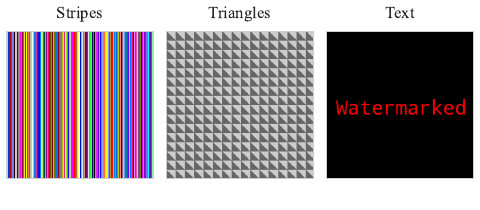}
\caption{\textbf{Visualizing the proposed triggers.} We visualize the proposed structured triggers. For \badnet-Stripes we use the ``Stripes'' trigger as a patch. For \blended-Stripes the ``Stripes'' trigger is overlayed on the full images with $n_c=0.03$ in~\cref{eqn:convex-sum}. ``Triangles'' and ``Text'' triggers are also overlayed on the original image as described in~\cref{sec:proposed-triggers}.}
\label{fig:triggers}
\end{figure}

\subsection{Cleaning methods}
\label{app:cleaning-methods}
For all (\cleanclip, \roclip, \ours) cleaning methods considered in this work, we optimize the hyper-parameters on a \badnet-Stripes poisoned RN50 and use these parameters across all other encoders and attacks. In this section we detail how hyper-parameter selection was done for different cleaning methods and related discussions.
\begin{table}[t]
\centering
\small
\tabcolsep=1.2pt
\extrarowheight=1.25pt
\newl=9mm
\newlc=11mm
\begin{tabular}{C{12mm} C{9mm} C{9mm} *{2}{|C{\newlc} C{\newlc}}}
\toprule
& & & \multicolumn{2}{c}{\imnet} & \multicolumn{2}{c}{MS-COCO} \\
\cline{4-7} 
Epochs & LR & BS & clean & ASR ($\downarrow$) & clean & ASR ($\downarrow$)  \\
\addlinespace[.8mm]
\toprule
5 & 1e-5 & 256 & 55.5 & 89.1 & 71.6 & 85.6\\
\rowcolor{NewGray} 5 & 2e-5 & 256 & 53.0 & 62.3 & 69.7 & 52.1 \\
5 & 3e-5 & 256 & 49.8 & 16.8 & 67.2 & 18.2 \\
\addlinespace[.8mm]
\midrule
5 & 2e-5 & 512 & 54.6 & 79.3 & 70.8 & 71.3 \\
\addlinespace[.8mm]
\midrule
10 & 2e-5 & 256 & 49.7 & 33.0 & 67.8 & 31.4\\
\bottomrule
\end{tabular}
\caption{\textbf{Optimizing \cleanclip.} Looking at \badnet-Stripes performance for different training setups for RN50. The selected parameters are \colorbox{NewGray}{highlighted}.} 
\label{tab:cleanclip-abl}
\end{table}

\begin{table}[b]
\centering
\small

%\vspace{2mm}
\tabcolsep=1.2pt
\extrarowheight=1.25pt
\newl=9mm
\newlc=11mm
\begin{tabular}{L{12mm} C{9mm} C{7mm} *{2}{|C{\newlc} C{\newlc}}}
\toprule
& & & \multicolumn{2}{c}{\imnet} & \multicolumn{2}{c}{MS-COCO} \\
\cline{4-7} 
Method & LR & $\mathcal{K}$ &clean & ASR ($\downarrow$) & clean & ASR ($\downarrow$)  \\
\toprule
\addlinespace[0.8mm]
\multicolumn{7}{l}{\textbf{Init:} Pre-trained \clip}\\
\midrule
\rowcolor{NewGray} \roclip & 5e-6 & 3 & 48.2 & 82.0 & 66.4 & 75.9\\
\roclip & 5e-6 & 2 & 47.9 & 83.2 & 66.9 & 69.9 \\
\roclip & 2e-5 & 3 & 40.7 & 45.0 & 60.7 & 45.9\\
% \toprule
% \addlinespace[0.8mm]
% \multicolumn{7}{l}{\textbf{Init:} Random, poison rate: 0.25\%}\\
% \midrule
% \roclip & 5e-5 & 2 & 2.1 & 21.8 & 5.7 & 25.7\\
\bottomrule
 \end{tabular}
 \caption{\textbf{Searching for optimal \roclip parameters.} Looking at \badnet-Stripes performance for different training setups for RN50 based \clip. Unlike~\cite{yang2024robust}, we start with a pre-trained model in accordance to our setup. The selected parameters are \colorbox{NewGray}{highlighted}.} 
\label{tab:roclip-abl}
\end{table}

\subsubsection*{\cleanclip}
Originally, \cleanclip uses a $BS$ of $64$ and fine-tune with $LR=1e-5$. Instead, we do a short sweep over LR, epochs and number of epochs in ~\cref{tab:cleanclip-abl}. We select the highlighted row with $\text{LR}=2e-5$ and $\text{BS}=256$ as it has better ASR 
%than higher LR 
while maintaining good clean performance. We use their default values\footnote{\href{https://github.com/nishadsinghi/CleanCLIP}{https://github.com/nishadsinghi/CleanCLIP}} for all other parameters (augmentations, loss weight, etc.) and train for 5 epochs. For every \cleanclip run with ResNet50 and \vit-B/32 based \clip, we stick to this setup. For \vit-L/14 based \clip runs, we reduce the $BS$ to 120 and train for 3 epochs while  adapting the schedule to 3 epochs. Reducing the number of epochs keeps the number of updates similar to the optimized schedule.

\myparagraph{Discussion regarding \cleanclip results in~\cite{liang2024badclip}.} For the \badclip attack against ResNet50-\clip, the authors in~\cite{liang2024badclip} optimized \cleanclip's hyper-parameters. Their parameters attain worse performance than what our optimized parameters get. Specifically, their run yields (clean, ASR) of (54.0\%, 89.6\%) while our found parameters (highlighted row in ~\cref{tab:cleanclip-abl}) yield much better (53.8\%, 40.1\%) performance. The clean accuracy we get is nearly the same as %\badclip's
in \cite{liang2024badclip} whereas we reduce ASR by more than 50\%. This justifies our chosen hyperparameters for \cleanclip.

\myparagraph{\cleanclip does not clean due to its loss.} In~\cref{tab:cleanclip-abl} for the $LR=3e-5$ setting, we show that \cleanclip can bring the ASR significantly down at the cost of clean performance. This happens as the higher $LR$ leads the model parameters to move away from the original poisoned model in some direction in the weight space and not because of the additional $\L_\text{UniAug}$ term in their $\L_\text{\cleanclip}$. If $\L_\text{UniAug}$ term had an effect on cleaning, it would show up in~\cref{fig:asr-ca-trade-off} where we have used the $\lambda$ values as high as 1000. Thus we infer that effective cleaning with \cleanclip is only possible by destroying the original model and not due to their proposed loss. On the contrary, $\L_\text{PAR}$ loss cleans by a structured deviation from the poisoned model while preserving clean performance that can be controlled by $\tau$.

\myparagraph{\cleanclip when structured trigger is added to its set of augmentations.}
In \cref{tab:reb-aug} we analyze if CleanCLIP can clean a backdoor attack if one adds the trigger (in this case ``stripes'') to the augmentation set. Indeed, in this case CleanCLIP works well even for the \blended-Stripes model which the standard version of CleanCLIP does not clean well. However, CleanCLIP still does not work for the other structured trigger \blended-Text. Thus there is no generalization across structured triggers. Moreover, it is clearly trivial to clean a model from a known backdoor trigger. In practice the attacker is free to use any backdoor trigger and thus this setting is not at all realistic. In contrast, \ours is independent of the structure of the backdoor trigger and works well across all employed triggers.
\begin{table}[h]
\centering
\small
\tabcolsep=1.2pt
\extrarowheight=-0.2pt
\newl=10mm
\newlc=14mm
\begin{tabular}{L{22mm} C{\newl} *{2}{|C{\newl} C{\newl}}}
& Stripes & \multicolumn{2}{c}{\imnet} & \multicolumn{2}{c}{COCO}  \\
\cline{3-6} 
Attack & Aug. &clean & ASR   & clean & ASR  \\
\toprule
\blended-Stripes & \xmark & 53.1 & 61.8 & 69.8 & 76.0\\
\blended-Stripes & \cmark & 53.3 & 0.1 & 70.2 & 1.4\\
\midrule
\blended-Text & \xmark & 53.3 & 42.4 & 70.2 & 32.5\\
\blended-Text &\cmark & 52.9 & 34.6 & 70.5 & 27.2\\
\bottomrule 
\end{tabular}
\vspace{-2mm}
\caption{\textbf{Stripes as augmentation in \cleanclip.} The \blended-Stripes model has a low ASR  when stripes is used as augmentation (Aug. \cmark) in comparison to the non-augmented version (Aug. \xmark).
}
% Moreover, this augmentation does not generalize to other structured patterns like \blended-Text.} 
\label{tab:reb-aug}
\end{table}
\subsubsection*{\roclip}

 Originally proposed for cleaning from scratch, we adapt \roclip~\cite{yang2024robust} to the fine-tuning setup. For this, we use their proposed\footnote{\href{https://github.com/BigML-CS-UCLA/RoCLIP}{https://github.com/BigML-CS-UCLA/RoCLIP}} $BS$, data augmentations, size of retrieval pool, learning rate schedule and number of epochs (24) trained for. We adapt their setup by fine-tuning with their method starting from a pre-trained \clip-RN50. As \roclip cleans with poisoned dataset, we keep the standard 0.5\% poisoning rate \ccm dataset ($2k/400k$) for training with it. Note, this means \roclip is effectively cleaning with an additional $150k$ samples in comparison to \cleanclip and \ours (as in this work, both of the latter use $250k$ samples to clean a poisoned model). 
 
 To select the optimal $LR$ and $\mathcal{K}$, the parameter that controls the frequency of epochs where the retrieval pool based loss is used, we do a sweep over some values in~\cref{tab:roclip-abl}. A higher $\mathcal{K}$ would increase clean performance at the cost of ASR. As $\mathcal{K}=3$ has 82\% ASR on \imnet, we do not increase it further. Inversely, a higher $LR$ cleans better (lower ASR)  at the cost of clean accuracy, hence we do not increase $LR$ beyond $2e-5$, as it already degraded clean accuracy a lot. Models trained with $LR < 5e-6$, yield models with very high ASR ($\sim85\%$) and we omit those values as well. Finally, we select the highlighted row as it achieves the best clean-ASR trade-off.
\subsection*{SafeCLIP}
\label{app:safeclip}
SafeCLIP~\cite{yangbetter} is a method for training \textit{from scratch} with a poisoned dataset, while PAR focuses on cleaning a backdoored \textit{pre-trained} CLIP.
We tried to adapt \safeclip to our finetuning setting which is non-trivial as the method is inherently a method for training from scratch. We could not achieve reasonable clean performance. For instance, applying \safeclip on a \blended-Stripes poisoned model, we get only 39\% clean accuracy on \imnet with ASR of 41\%, in comparison to 53.5\% clean and 0.1\% ASR with \ours. Using \safeclip poisoned images (poisoned models were not available but the authors provided us kindly with the poisoned training images),  we trained a model from scratch in their setup (1M samples from CC3M, 0.05\% poisoning rate) with \blended. This poisoned model yields a clean performance on \imnet of 5.9\% and has an ASR of 91.3\%. On cleaning this model with the default \ours setup, we could completely get rid of backdoors with an ASR of 0.0\% and even increase slightly the clean performance to \textbf{6.1\%}. This showcases the effectiveness of \ours against models poisoned from scratch. Thus \ours can always be applied after training a (poisoned) model from scratch while we could not adapt \safeclip for cleaning via finetuning (in contrast to \roclip).

\begin{table}[t]
\centering

\small
\tabcolsep=1.1pt
\extrarowheight=2pt
\begin{tabular}{L{5mm} L{25mm} | C{25mm}  | C{25mm}  }
& Configuration & RN50/\vit-B/32 & \vit-L/14  \\
% \cline{3-4} 
% & & PSPNet & \upernet & \upernet & Segmenter \\
\toprule
\parbox[t]{4mm}{\multirow{6}{*}{\rotatebox[origin=c]{90}{\textcolor{cvprblue}{DATA}}}} & 
Data    &  \ccm/\synth & \ccm  \\
& Image size    &  224x224 & 336x336  \\
\cline{2-4}
& \multirow{2}{*}{Gaussian Noise}  &  Std = 0.2 & Std = 0.2  \\
& &Prob = 0.5 & Prob = 0.5 \\
\cline{2-4}
& \multirow{2}{*}{CutOut-Patch}  &  Area = (0.5 - 1)\% & Area = (0.5 - 1)\%  \\
& & Prob = 0.5 & Prob = 0.5 \\
\cline{2-4}
\parbox[t]{4mm}{\multirow{10}{*}{\rotatebox[origin=c]{90}{\textcolor{cvprblue}{TRAINING \& LOSS}}}} & Threshold ($\tau$) & 2.15 & 2.15 \\
& Optimizer    &  AdamW & AdamW  \\
& Start-LR  &  3e-5 & 3e-5 \\ 
& Peak-LR  &  3e-6 & 3e-6 \\ 
& Final-LR  &  1e-9 & 1e-9 \\ 
& Peak-epoch  &  5 & 2.5 \\ 
& Epochs  &  10 & 5 \\ 
& Weight decay   &  1e-4 & 1e-4 \\ 
& Batch size   &  512x4 & 120x8 \\ 
& Momentum & 0.9, 0.999 & 0.9, 0.999  \\
\bottomrule
\end{tabular} 
\caption{\textbf{%
Training and data configurations.} The specific training settings for different models trained in this work using \ours. We differentiate between small encoders in \clip like RN50/\vit-B and the large one in \vit-L/14.}
\label{tab:train-config}
\end{table}

\subsubsection*{Perturb and Recover (\ours)}
We use a custom learning schedule as highlighted in the main part. We emphasize that for the standard cosine decayed LR schedule as used by \cleanclip, our method still outperforms \cleanclip, see~\cref{tab:our-othersched}. All other details specific to \ours can be found in~\cref{tab:train-config}. As it can be noted from~\cref{tab:train-config}, we train for half the number of epochs for \vit-L/14 as the BS is halved and this keeps the effective number of updates of the model similar. Even though we fix the threshold $\tau$ to 2.15 for all our experiments, one can choose the optimal threshold to control the clean accuracy-ASR trade-off with our method, using curves like the one in~\cref{fig:asr-ca-trade-off}. As we have two data-augmentations applied with a certain probability in \ours, we look at the variance across a few attacks in~\cref{tab:variance}. The variance is marginal across attacks and evaluated metrics.

\subsubsection*{Computational Costs}
We define one unit of computation as a complete forward and backward pass through both image and text encoders. Our method, \ours requires 1.5 units of computation, consisting of one full forward and backward pass (1 unit) plus an additional forward pass through the frozen poisoned encoders (0.5 units). Similarly, \cleanclip also requires 1.5 units, comprised of one full forward and backward pass (1 unit) plus one additional forward pass with augmented image and text data (0.5 units). In contrast, standard CLIP training uses only one forward and one backward pass, totaling 1 unit of computation. Therefore, both \ours and \cleanclip have identical computational requirements, which amount to 1.5 times the cost of standard \clip training. \roclip and \safeclip are both methods designed for training from scratch and partially involve more complex steps so that it is difficult to compare their cost to \cleanclip and \ours.
%\roclip is a method for training from scratch, hence trained for much longer than \ours and \cleanclip, making it's comparison here unjust.

\section{Additional Experiments}
\label{app:additional-exp}

In this section, we expand on additional experiments that highlight the effectiveness of both the proposed triggers and backdoor removal method \ours. 
\subsection{Effectiveness of the proposed triggers}
\textbf{Bypassing \cleanclip. } In~\cref{tab:structured-patterns}, we replace the standard random noise in the \blended attacks with the proposed triggers: Stripes, Triangles and Text. For both RN50 and \vit-B/32 based \clip, the proposed triggers are as effective as random noise in terms of backdoor attack success rate (ASR).  While \cleanclip reduces ASR for the original random pattern very well, for the structured patterns the ASR is in most cases still very high. This shows how the proposed triggers are more effective against strong augmentation based cleaning methods like \cleanclip than random noise. In~\cref{fig:vis-backdoor-app}, we show more visualizations with the proposed triggers. 

The effectiveness of the proposed triggers can also be visualized via t-SNE projections of the embedding of the image encoder in trained \clip models in the ``\textsc{t-sne embeddings}'' blocks in Figures~\ref{fig:tsne-badnet-stripes} and~\ref{fig:train-tsne-blendrs}. For the poisoned \clip the embeddings of original and backdoored images form disjoint clusters, indicating the model differentiates among these. After~\cleanclip, the embeddings in Figures~\ref{fig:tsne-badnet-stripes} and~\ref{fig:train-tsne-blendrs} are still disjoint. But, the same \cleanclip embeddings for the random noise based~\badnet in~\cref{fig:train-tsne-badnet-random} are homogeneously distributed. This clearly points to the fact that random noised trigger based attacks are easily cleaned by methods like \cleanclip, which is not the case when one transitions to structured trigger based attacks.

\begin{table}[!b]
\centering
\small

%\vspace{2mm}
\tabcolsep=1.1pt
\extrarowheight=1.25pt
\newl=9mm
\newlc=11mm

\begin{tabular}{L{12mm} C{8mm} C{12mm} *{2}{|C{\newlc} C{\newlc}}}
\toprule
& & & \multicolumn{2}{c}{\clip} & \multicolumn{2}{c}{\cleanclip} \\ %& \multicolumn{2}{|c}{\ours} \\
\cline{4-7} 
Attack.  & $n_c$ & Source & clean & ASR ($\uparrow$)  & clean & ASR ($\uparrow$)  \\ % & clean & ASR($\downarrow$) \\
\toprule
\addlinespace[1.1mm]
\multicolumn{5}{l}{\textbf{\clip encoder:} ResNet-50} \\
\midrule
Random & 0.2 &\cite{chen2017targeted} & 57.7 & 99.4 & 53.4 & 19.5 \\ %& 53.7 & 0.0\\
\addlinespace[0.5mm]
\cdashline{1-7}
\addlinespace[0.5mm]
Stripes & 0.03 & ours & 57.6 & 95.6 & 53.1 & 61.8\\ % & 54.1 & 1.0 \\
Triangles & 0.15 & ours & 57.4 & 85.7 & 52.9 & 48.7\\ % & 53.0 & 22.0 \\
Text & 0.5 & ours & 56.9 & 95.6 & 53.2 & 42.4 \\ %& 53.7 & 22.1 \\
\toprule
\addlinespace[1.1mm]
\multicolumn{5}{l}{\textbf{\clip encoder:} ViT-B/32} \\
\midrule
Random & 0.2 &\cite{chen2017targeted}& 59.3 & 51.5 & 54.3 & 0.0 \\
\addlinespace[0.5mm]
\cdashline{1-7}
\addlinespace[0.5mm]
Stripes &0.03 & ours & 59.7 & 99.8 & 55.0 & 15.2  \\
Triangles & 0.15 & ours& 58.9 & 99.7 &  54.6 & 91.4 \\
Text & 0.5 &ours & 59.3 & 99.8 &  54.7 & 62.9 \\
\bottomrule
\end{tabular}
\caption{\textbf{Structured trigger patterns are more effective than random noise.} For the \blended~\cite{chen2017targeted} attack model, we replace the original random noise with our structured patterns. We evaluate for both RN50 and \vit-B/32 based \clip for the zero-shot classification for \imnet.} 
\label{tab:structured-patterns}
\end{table}

\myparagraph{Other target labels.} In~\cref{tab:other-labels}, we show for RN50 based \clip the proposed triggers work across different target labels. Specifically, in addition to the target ``banana'', we show for \imnet classes ``refrigerator'' and ``strawberry'' \badnet-Stripes attains high ASR while poisoning \clip and bypasses \cleanclip as well. Note, for the label ``strawberry'' we cannot provide an ASR for COCO retrieval task as there are no captions with ``strawberry'' in the validation set we use.

\myparagraph{Lower poisoning rate.} In the pink bars in~\cref{fig:poison-rate}, we plot the ASR as attained by \badnet-Stripes across a range of poisoning rate. Even at a meager poisoning rate of 0.05\%, \badnet-Stripes achieves an impressive 92\% success rate. On increasing the poisoning rate, the ASR very quickly goes to 99\%.

\begin{table}[!t]
\centering
\small
\tabcolsep=1.2pt
\extrarowheight=1.25pt
\newl=9mm
\newlc=12mm
\begin{tabular}{L{20mm}  *{2}{|C{\newlc} C{\newlc}}}
\toprule
& \multicolumn{2}{c}{\imnet} & \multicolumn{2}{c}{MS-COCO} \\
\cline{2-5} 
Method & clean & ASR ($\downarrow$) & clean & ASR ($\downarrow$)  \\
\toprule
\addlinespace[0.8mm]
\multicolumn{5}{l}{\textbf{Target:} banana}\\
\midrule
\rowcolor{NewGray}\clip &  57.6 & 99.8 & 73.1 & 99.9 \\
\cleanclip & 53.0 & 62.3 & 69.7 & 52.1 \\
\ours  & 53.0 & \textbf{42.4} & 70.2 & \textbf{20.9} \\
\midrule
\addlinespace[0.8mm]
\multicolumn{5}{l}{\textbf{Target:} refrigerator}\\
\midrule
\rowcolor{NewGray}\clip & 57.5 & 99.2 & 73.2 & 98.8 \\ 
\cleanclip & 53.4 & 37.7 & 69.9 & 25.9 \\
\ours & 54.0 & \textbf{34.4} & 72.3 & \textbf{13.8} \\
\midrule
\addlinespace[0.8mm]
\multicolumn{5}{l}{\textbf{Target:} strawberry}\\
\midrule
\rowcolor{NewGray}\clip  & 57.6 & 99.2 & 72.9 & - \\ 
\cleanclip  & 53.6& 72.9 & 69.6& - \\
\ours  & 54.0 & \textbf{28.8} & 71.4 & - \\
\bottomrule
\end{tabular}
\caption{\textbf{Proposed \badnet-Stripes and \ours work across target labels.} For the target ``strawberry'' we can't provide an ASR for COCO as ``strawberry'' is not present in any of the $5k$ validation samples we use. The original poisoned \clip model is \colorbox{NewGray}{highlighted}.} 
\label{tab:other-labels}
\end{table}

\begin{table}[b]
\centering
\small

%\vspace{2mm}
\tabcolsep=1.2pt
\extrarowheight=1.25pt
\newl=9mm
\newlc=14mm
\begin{tabular}{L{22mm} *{2}{|C{\newlc} C{\newlc}}}
\toprule
& \multicolumn{2}{c}{\imnet} & \multicolumn{2}{c}{COCO}  \\
\cline{2-5} 
Attack & clean & ASR ($\downarrow$)  & clean & ASR ($\downarrow$) \\
\toprule
\blended-Stripes & 53.5\small{$ \pm $0.2} & 0.1\small{$ \pm $0.1} & 70.7\small{$ \pm $0.1} & 5.0\small{$ \pm $0.4}\\
\badnet-Stripes & 53.0\small{$ \pm $0.6} & 42.4\small{$ \pm $6.1} & 70.2\small{$ \pm $0.2} & 20.9\small{$ \pm $3.2} \\
\badclip & 53.4\small{$ \pm $0.4} & 30.4\small{$ \pm $3.1} & 70.7\small{$ \pm $0.6} & 31.2\small{$ \pm $2.4}\\
\bottomrule
 
\end{tabular}
\caption{\textbf{Variance of \ours.} We compute the mean and standard deviations across 3 runs for \ours. Overall the variance across runs and attacks is marginal.} 
\label{tab:variance}
\end{table}

\subsection{Backdoor removal with \ours}
\label{app:extra-par}
\textbf{Effectiveness at lower poisoning rates.} As prior works~\cite{liang2024badclip, bansal2023cleanclip} use a poisoning rate of 0.3\%, we plot in~\cref{fig:poison-rate} ASR attained at poisoning rate as low as 0.05\%: across all tested poisoning rates, \ours outperforms \cleanclip. Specifically, at 0.25\% poisoning rate \cleanclip has a ASR of 40\% whereas \ours attains a significantly lower ASR of 19\%.

\myparagraph{\ours works for general LR schedules.} As mentioned in the main part, to speed up cleaning with \ours we use a custom learning schedule. To disentangle the effect of this schedule from the novel loss (\cref{eqn:loss}) term we propose, we clean with $\L_\text{\ours}$ in the cosine schedule of~\cleanclip and present the results in~\cref{tab:our-othersched}. For both \badnet-Stripes and~\blended-Stripes, \ours cleans way better than \cleanclip. For~\blended-Stripes, \ours even achieves a 2\% clean accuracy improvement over \cleanclip.

\begin{table}[b]
\centering
\small
%\vspace{2mm}
\tabcolsep=1.2pt
\extrarowheight=1.25pt
\newl=9mm
\newlc=12mm
\begin{tabular}{L{22mm} *{2}{|C{\newlc} C{\newlc}}}
\toprule
& \multicolumn{2}{c}{\imnet} & \multicolumn{2}{c}{MS-COCO} \\
\cline{2-5} 
Method & clean & ASR ($\downarrow$) & clean & ASR ($\downarrow$)  \\
\toprule
\addlinespace[0.8mm]
\multicolumn{5}{l}{\textbf{Attack:} \badnet-Stripes}\\
\midrule
\cleanclip  &53.0 & 62.3 & 69.7 & 52.1\\
\ours & 53.1& \textbf{51.5} &  69.3 &  \textbf{29.8}\\
\toprule
\addlinespace[0.8mm]
\multicolumn{5}{l}{\textbf{Attack:} Blended-Stripes}\\
\midrule
\cleanclip & 53.1 & 61.8 & 69.8 & 76.0\\
\ours & 55.1 & \textbf{25.4} &70.3 & \textbf{17.4}\\
\bottomrule
\end{tabular}
\caption{\textbf{Learning rate schedule has little effect on performance.} We train with \ours using the schedule from \cleanclip with peak-LR set to 2e-5.} 
\label{tab:our-othersched}
\end{table}

\begin{table}[t]
\centering
\small
%\vspace{2mm}
\tabcolsep=1.2pt
\extrarowheight=1.25pt
\newl=9mm
\newlc=11mm
\begin{tabular}{L{20mm} C{8mm} *{2}{|C{\newlc} C{\newlc}}}
\toprule
& & \multicolumn{2}{c}{\imnet} & \multicolumn{2}{c}{MS-COCO} \\
\cline{3-6} 
Method & $\tau$ & clean & ASR($\downarrow$) & clean & ASR($\downarrow$)  \\
\toprule
\addlinespace[0.8mm]
\multicolumn{6}{l}{\textbf{Attack:} \badnet-Stripes}\\
\addlinespace[0.8mm]
\cline{1-6}
\addlinespace[0.8mm]
\cleanclip  & --& 54.7 & 86.8 & 70.5 & 84.7\\
\ours & 2.15 & 54.4 & 50.1 & 69.5 & 53.5 \\
\ours & 2.25 & 53.4 & 44.2 & 68.6 & 48.8 \\
 \ours & 2.3 & 52.3 & 40.6 & 65.0 & 42.3 \\
\addlinespace[0.8mm]
\cline{1-6}
\addlinespace[0.8mm]
\multicolumn{6}{l}{\textbf{Attack:} Blended-Triangles}\\
\addlinespace[0.8mm]
\cline{1-6}
\addlinespace[0.8mm]
\cleanclip  & -- & 54.6 & 91.4 & 70.2 & 98.5\\
\ours & 2.15 & 54.9 & 15.6 & 70.2 & 91.5 \\
\ours & 2.25 & 54.4 & 14.7 & 68.3 & 94.0 \\
\ours & 2.3 & 53.8 & 12.1 & 67.6 & 89.2\\
% & \multicolumn{6}{l}{\textbf{Attack:} \badnet-Stripes}\\
% \cline{2-7}
% \addlinespace[0.8mm]
% \multirow{3}{*}{\rotatebox{90}{\textcolor{cvprblue}{\vit-L/14}}} & \cleanclip  & -- & 68.7 & 84.5 & 78.8 & 6.3\\
% & \ours & 2.15 & 65.8 & 0.0 & 76.7 & 2.1\\
% & \ours & 2.0 &  & 0.1 & \\
% \addlinespace[0.8mm]
% \addlinespace[0.8mm]
% \multicolumn{6}{l}{\textbf{Attack:} \badclip}\\
% \midrule
% \cleanclip  & -- & 55.2 & 19.6 & 70.1 & 28.5\\
% \ours & 2.15 & 54.7 & 18.2 & 70.6 & 27.1 \\
% \ours & 2.25 & 53.8 & 16.1 & 68.8 &  25.2\\
\bottomrule
\end{tabular}
\caption{\textbf{Sweeping over $\tau$ for \vit-B/32 for clean-ASR trade-off.} Numbers are shown for \ours with the fixed value of $2.15$ in the main part and one more value for the two architectures each. For easier comparison, \cleanclip results are also illustrated.}
\label{tab:tau-vit}
\end{table}

\begin{table}[t]
\centering
\small
\tabcolsep=0.7pt
\extrarowheight=1.5pt
\newl=9mm
\newlc=11mm
\begin{tabular}{L{16mm} C{17mm} *{2}{|C{\newlc} C{\newlc}}}
\toprule
% & & \multicolumn{2}{c|}{\badnet} & \multicolumn{2}{c}{\blended} 
% \\
 & & \multicolumn{2}{c|}{\badnet-Stripes}  & \multicolumn{2}{c}{\blended-Text}  \\
\cmidrule{3-6}
Eval Data & Method & clean & ASR  & clean & ASR \\ % & clean & ASR  \\
\addlinespace[0.5mm]
\toprule 
\addlinespace[0.5mm]
% \rowcolor{NewGray} \cellcolor{white} \multirow{2}{*}[-0.4em]{\rotatebox{90}{\textcolor{cvprblue}{\imnet}}}
% \rowcolor{NewGray}\imnet & \clip & 75.0 & 99.9 & 74.8 & 92.3  \\
% \addlinespace[0.5mm]
% \cdashline{2-6}
% \addlinespace[0.5mm]
% \imnet & \cleanclip & 70.3 & 15.6 & 65.1 & 22.7 \\
% \imnet & \ours (ours) & 64.4 & 0.1 & \textbf{65.6} & \textbf{8.1} \\
% \addlinespace[0.5mm]
% \midrule
% \addlinespace[0.5mm]
% % \rowcolor{NewGray} \cellcolor{white} \multirow{2}{*}[-0.2em]{\rotatebox{90}{\textcolor{cvprblue}{COCO}}} 
% \rowcolor{NewGray}COCO & \clip & 80.2 & 99.8 & 79.6 & 45.8\\
% \addlinespace[0.5mm]
% \cdashline{2-6}
% \addlinespace[0.5mm]
% COCO & \cleanclip & 78.2 & 29.4 & \textbf{75.6} & 14.2   \\
% COCO & \ours (ours) & 72.6 & 1.6 & 73.6 &  \textbf{6.5}\\
\rowcolor{NewGray}\imnet & \clip & 73.5 & 100.0 & 73.7 & 99.3  \\
\addlinespace[0.5mm]
\cdashline{2-6}
\addlinespace[0.5mm]
\imnet & \cleanclip & 68.7 & 84.5 & 68.9 & 83.4\\
\imnet & \ours (ours) & 65.8 & \textbf{0.0} & 65.2 & \textbf{17.0}\\
\addlinespace[0.5mm]
\midrule
\addlinespace[0.5mm]
% \rowcolor{NewGray} \cellcolor{white} \multirow{2}{*}[-0.2em]{\rotatebox{90}{\textcolor{cvprblue}{COCO}}} 
\rowcolor{NewGray}COCO & \clip & 80.5 & 22.6 & 79.9 & 97.7 \\
\addlinespace[0.5mm]
\cdashline{2-6}
\addlinespace[0.5mm]
COCO & \cleanclip & 78.8 & 6.3 & 78.7 & 65.1 \\
COCO & \ours (ours) & 76.7 & \textbf{2.1} & 76.0 & \textbf{10.2} \\
\bottomrule
\end{tabular}
\caption{\textbf{Testing the attack and defense on \vit-L/14-336.} We scale \ours to the larger \vit-L/14. The poisoning and cleaning is done at the higher resolution of $336 \times 336$.} 
\label{tab:vitl-exp}
\end{table}

\begin{table}[!t]
\centering
\small
\tabcolsep=1.0pt
\extrarowheight=0.5pt
\newl=10mm
\newlc=16mm
\begin{tabular}{L{20mm} C{\newlc} *{2}{|C{\newl} C{\newl}}}
\toprule
& & \multicolumn{2}{c}{\imnet} & \multicolumn{2}{c}{COCO}  \\
\cline{3-6} 
Attack & Model &clean & ASR   & clean & ASR  \\
\toprule
\rowcolor{NewGray}\blended-Text & Poisoned & 74.7 & 99.4 & 85.1 & 99.5\\
\blended-Text & \cleanclip & 68.4 & 87.6 & 79.9 & 90.1\\
\blended-Text & \ours & 69.3 & \textbf{30.9} & 72.7 & \textbf{60.4}\\
\midrule
\rowcolor{NewGray}\badnet-Stripes & Poisoned & 74.8 & 99.8 & 85.0 & 97.8\\
\badnet-Stripes & \cleanclip & 70.2 & 56.8 & 80.4 & 61.8\\
\badnet-Stripes & \ours & 70.1 & \textbf{36.6} & 74.4 &  \textbf{60.1}\\
\bottomrule 
\end{tabular}
\vspace{-2mm}
\caption{\textbf{Other models.} We poison \textbf{SigLIP} (ViT-B/16) model with the same setup as \clip models. Due to the loss of SigLIP being at a different scale, we select a higher $\tau$ of 3.4 (without grid-search). The original poisoned SigLip model is \colorbox{NewGray}{highlighted}.}
\vspace{-5mm}
\label{tab:reb-siglip}
\end{table}

\myparagraph{Scaling \ours to \vit-L/14.}
In~\cref{tab:vitl-exp}, we show the effectiveness of \ours for the \vit-L/14 based \clip model. Specifically, we poison the model with \badnet-Stripes and \blended-Text attacks at a higher resolution of $336 \times 336$. As mentioned in~\cref{app:exp-details}, we poisoned the models for 3 additional epochs (8 epochs in total) in comparison to \vit-B poisoning as achieving ASR $> 95\%$ on \imnet for both attacks was not possible with 5 epochs. In this setting as well, \ours significantly outperforms \cleanclip with a marginal degradation in clean performance. This result is very promising as larger resolutions of $336\times 336$ pixels are often used in downstream LVLMs \eg LLaVa~\cite{liu2024improved} and VILA~\cite{lin2024vila}. However, our threat model of poisoning \clip models with backdoored images-caption pairs does not allow a simple transfer to LVLMs as these models only use the vision encoder of the \clip models. Without the text encoder, the backdoor is not effective. 

\myparagraph{\ours for SigLIP.}
In~\cref{tab:reb-siglip}, we show the effectiveness of \ours for the ViT-B/16 SigLIP model. We poison the model with \badnet-Stripes and \blended-Text attacks. As the loss of SigLIP is working on a different scale, we select the larger $\tau$ of 3.4 without employing a grid search as before which is likely suboptimal. \ours outperforms CleanCLIP also for this other type of CLIP models with improvements of 56.7\% in ASR for Blended-Text and 20.2\% in ASR for BadNet-Stripes for zero-shot ImageNet classification while maintaining higher clean accuracy.
 
\myparagraph{Trade-off for \vit-B/32.} While in~\cref{fig:asr-ca-trade-off} we plot the clean accuracy and ASR trade-off achieved by $\tau$ parameter in \ours for ResNet50 based \clip, one can do the same to select the optimal $\tau$ for other models. In~\cref{tab:tau-vit} we evaluate for an additional value of $\tau$ against a subset of attacks. As expected on increasing $\tau$ from $2.15$ to $2.25$ we further reduce the ASR with a slight degradation in clean performance. This points to the fact that one would get a trade-off plot similar to the one for ResNet50 cleaned by \ours.

\myparagraph{If clean data has some backdoored samples?.}
We test what happens if the clean data has small amount of backdoored samples in it. For this, we add 50 \blended-Stripes poisoned images to our 250k fine-tuning set (RN50 poisoned model). Again, \ours shows effective cleaning: we get ASR for \imnet down from 95.6\% to 10.7\%  with clean performance of 50.8\% and ASR for retrieval down from 98.2\% to 11.4\%.

\myparagraph{\ours for a non-backdoored model.}
To test how \ours works when the model does not have any backdoor trigger, we fine-tune a clean (non-poisoned) \clip RN50 with \ours and get 54.3\% and 70.0\% clean accuracy on ImageNet and COCO respectively, compared to the clean model's 59.6\% and 72.8\%. Such drop is similar to what observed when cleaning poisoned models, see~\cref{tab:rn50-novel}, and can be controlled by adjusting $\tau$ in~\cref{fig:asr-ca-trade-off}.
\begin{table}[b]
\centering
\small
\tabcolsep=1.2pt
\extrarowheight=2pt
\newl=10mm
\newlc=14mm
\begin{tabular}{%L{12mm}
L{35mm} *{2}{|C{\newl} C{\newl}}}
%&  
\toprule
& \multicolumn{2}{c}{\imnet} & \multicolumn{2}{c}{COCO}  \\
\cline{2-5} 
%Method & 
Augmentations in PAR & clean & ASR   & clean & ASR  \\
\toprule
CutOut & 52.4 & 18.7 & 66.8 & 12.4\\

Gaussian Noise & 50.9 & 0.2 & 67.3 & 4.2 \\
CutOut + Uniform Noise & 50.7 &  1.0 & 66.6 & \textbf{4.0}\\
CutOut + Gaussian Noise & \textbf{53.5} & \textbf{0.1} & \textbf{70.7} & 5.0\\
\bottomrule 
\end{tabular}
\caption{\textbf{Performance of \ours varying augmentations.} Using just a single augmentation is enough for \ours to work, but a combination of CutOut + Gaussian Noise, yields the best backdoor removal/clean performance tradeoff.}
\label{tab:reb-aug}
\end{table}

\myparagraph{Effect of data augmentations used in \ours.} 
We report the performance of PAR with different augmentations in Table~\ref{tab:reb-aug} (RN50, \blended-Stripes attack). While using a single augmentation already leads to good accuracy and backdoor removal, combining methods yields further improvements. From this ablation, we see that the selected augmentations with CutOut and Gaussian noise achieves the best backdoor removal-performance tradeoff.

\myparagraph{Effect of number of clean real samples on \ours.}
In ~\cref{tab:less-clean-data} we reduce the clean training samples from \ccm: as expected, more data helps both accuracy and backdoor removal, PAR is already effective with as few as 100k samples for \badnet and 50k samples for the \blended attacks.
\begin{table}[t]
\centering
\small
\tabcolsep=1.1pt
\extrarowheight=1pt
\newl=10mm
\newlc=14mm
\begin{tabular}{L{15mm} L{8mm} C{14mm} *{2}{|C{\newl} C{\newl}}}
\toprule
& & & \multicolumn{2}{c}{\imnet} & \multicolumn{2}{c}{COCO}  \\
\cline{4-7} 
Attack & Model & Samples & clean & ASR   & clean & ASR  \\
\toprule
\multirow{4}{*}{\makecell[l]{\badnet-\\Stripes}}& \clip & - & 57.6 & 99.8 & 73.1 & 99.9\\
& PAR & 50k & 48.4 & 76.8 & 69.3 & 64.4\\
& PAR & 100k & 48.8 & 56.0 & 66.6 & 22.0\\
& PAR & 250k & 53.0 & 42.4 & 70.2 & 20.9\\
\midrule
\multirow{4}{*}{\makecell[l]{\blended-\\Stripes}}& \clip & - & 57.6 & 95.6 & 73.2 & 98.2\\
& PAR & 50k & 48.4 & 13.4 & 68.7 & 7.0\\
& PAR & 100k & 48.1 & 0.5 & 65.6 & 4.0\\
& PAR & 250k & 53.5 &0.1 & 70.7 & 5.0\\
\bottomrule 
\end{tabular}
\caption{\textbf{Performance of \ours varying clean training samples.} \ours is still effective when having access to as low as 50k clean samples across different attacks.}
\label{tab:less-clean-data}
\end{table}
\subsection{Evaluating COCO retrieval}
For COCO retrieval, until now we focused on top-5 clean text retrieval. In~\cref{tab:rn50-cocoretrieval} and~\cref{tab:vitb-cocoretrieval}, we show how \cleanclip and \ours fare against different backdoor attacks for standard top-1 image and text retrieval for RN50 and ViT-B/32. Across all attacks for ResNet50 based \clip, \ours achieves a higher retrieval rate when the inputs are backdoored (``bkd'' column in~\cref{tab:rn50-cocoretrieval}). The clean image retrieval performance of \ours is marginally lower than \cleanclip on average whereas text retrieval is always better. Overall the drop in clean performance from the original poisoned \clip (highlighted row in table) to \ours is marginal. In general, a similar trend holds for \vit-B/32 in~\cref{tab:vitb-cocoretrieval}.

\section{More Visualizations}
\label{app:visualize}
\begin{figure*}[t]
\centering
    \includegraphics[width=0.95\linewidth]{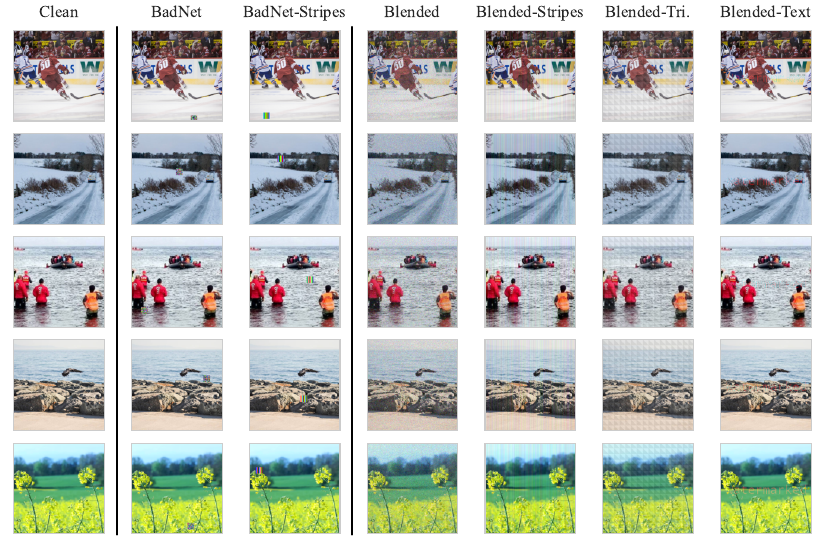}
\caption{\textbf{Visualizing more images with known and proposed triggers.} Standard \badnet~\cite{gu2017badnets} and \blended~\cite{chen2017targeted} use Gaussian noise as a trigger, we replace the noise with random stripped pattern for \badnet termed \badnet-Stripes. For the \blended attack, we further replace the random noise with stripes, low contrast triangles (Blended-Triangles) and ``Watermarked'' text (\blended-Text). \textit{Note: this is a very small subset of possible structured patterns, and we believe similar other patterns would be equally effective}.}
\label{fig:vis-backdoor-app}
\vspace{1cm}
  \includegraphics[width=0.95\linewidth]{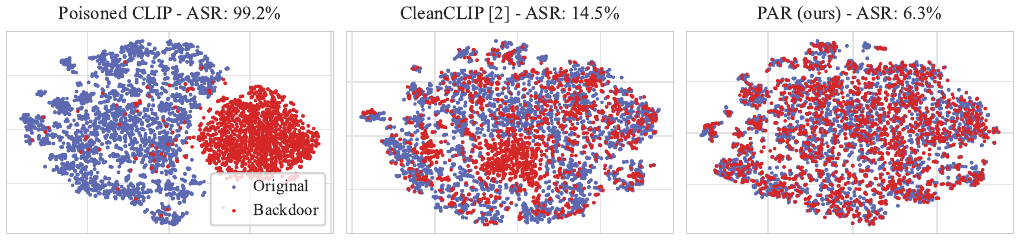}
\caption{\textbf{Visualizing the embeddings of different models for \badnet~\cite{gu2017badnets} poisoned RN50.} We visualize the t-SNE projections of random noise based \badnet poisoned \clip, clean fine-tuned by \cleanclip and fine-tuned by \ours. In this case, \cleanclip embeddings are much more homogeneously distributed in comparison to the proposed attacks with structured patterns. This shows that for random noise based triggers, \cleanclip can be effective. Overall the spread of points achieved by \ours is most homogeneous which reflects in the lowest ASR.}
\label{fig:train-tsne-badnet-random}
\end{figure*}

\begin{figure*}[t]
\small \centering
\begin{tikzpicture}
      % \node[anchor=north west, fill=blue!10] at (0.8, 0) {%
 \node[anchor=north west, minimum height=2.0cm, minimum width=0.98\linewidth, rounded corners=8pt] at (0.2, 0) { };
\node[anchor=north west] at (1.0, -0.0) {\includegraphics[width=0.95\linewidth]{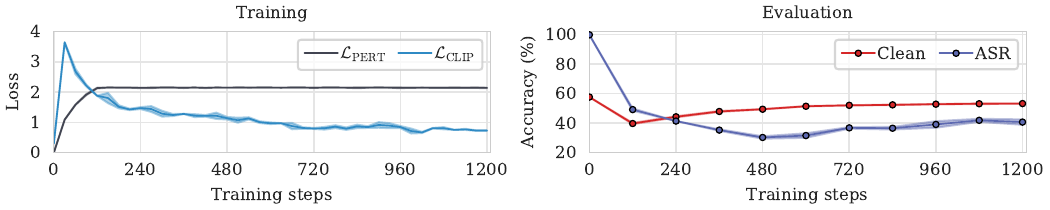} };   
\node[rotate=90, anchor=center] at (0.5, -1.6) { \textsc{Training \ours}};     
\node[anchor=north west, minimum height=4.6cm, minimum width=0.98\linewidth, rounded corners=8pt] at (0.2, -2) { };
\node[anchor=north west] at (1.2, -3.8) {\includegraphics[width=0.93\linewidth]{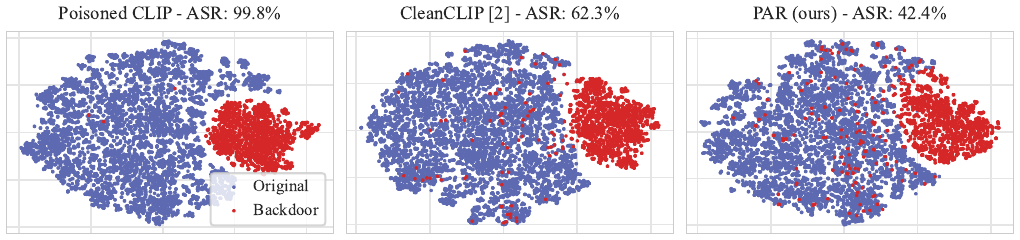} };   
\node[rotate=90, anchor=center] at (0.55, -6) {\textsc{t-SNE embeddings}}; 
\end{tikzpicture}
\vspace{-0.1cm}
\caption{\textbf{Training dynamics and visualizing the embeddings of different models for \badnet-Stripes poisoned RN50.} In the top left plot, we show how the $\L_{\text{CLIP}}$ and $\L_{\text{PERT}}$ ($\tau=2.15$) loss terms develop over training steps (evaluated every 25 steps) for \badnet-Stripes poisoned RN50. In the top right plot, we see how the training schedule generalizes by plotting clean accuracy and ASR (evaluated on $10k$ samples from \imnet). In the bottom row, we visualize the t-SNE projections of the same \badnet-Stripes poisoned \clip, clean finetuned by \cleanclip and finetuned by \ours. Overall \ours yields the best mixing of clean and backdoored samples. Better mix means the model sees the clean and backdoored samples similarly, which also translates to low ASR.}
\label{fig:tsne-badnet-stripes}
\end{figure*}

\begin{figure*}[b]
\small \centering
\begin{tikzpicture}
      % \node[anchor=north west, fill=blue!10] at (0.8, 0) {%
 \node[anchor=north west, minimum height=2.0cm, minimum width=0.98\linewidth, rounded corners=8pt] at (0.2, 0) { };
\node[anchor=north west] at (1.0, -0.0) {\includegraphics[width=0.95\linewidth]{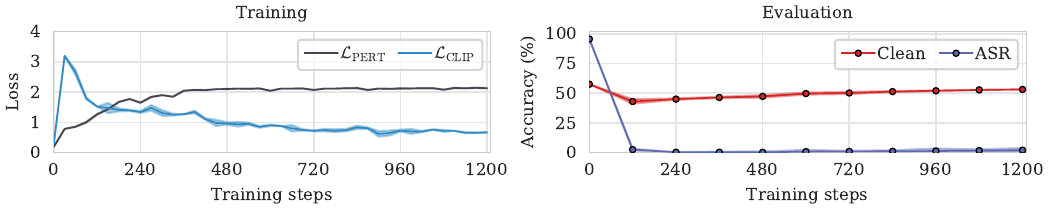} };   
\node[rotate=90, anchor=center] at (0.5, -1.6) { \textsc{Training \ours}};     
\node[anchor=north west, minimum height=4.6cm, minimum width=0.98\linewidth, rounded corners=8pt] at (0.2, -2) { };
\node[anchor=north west] at (1.2, -3.8) {\includegraphics[width=0.93\linewidth]{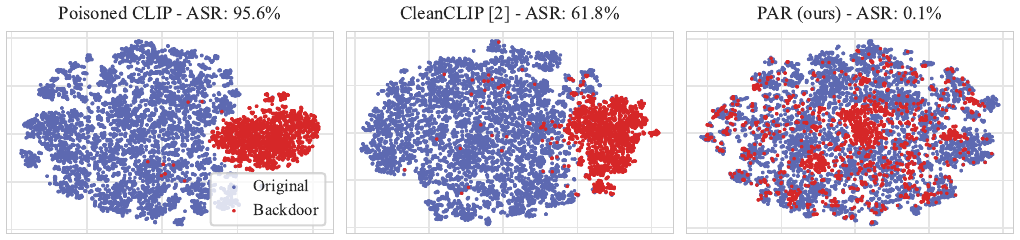} };   
\node[rotate=90, anchor=center] at (0.55, -6) {\textsc{t-SNE embeddings}}; 
\end{tikzpicture}
\vspace{-0.1cm}
\caption{\textbf{Training dynamics and visualizing the embeddings of different models for \blended-Stripes poisoned RN50.} In the top left plot, we show how the $\L_{\text{CLIP}}$ and $\L_{\text{PERT}}$ ($\tau=2.15$) loss terms develop over training steps (evaluated every 25 steps) for \blended-Stripes poisoned RN50. Even though the schedule was optimized for \badnet-Stripes poisoned RN50, in the top right plot, we see how the training schedule generalizes by plotting clean accuracy and ASR (evaluated on $10k$ samples from \imnet). In the bottom row, we visualize the t-SNE projections of the same \blended-Stripes poisoned \clip, clean finetuned by \cleanclip and finetuned by \ours. Overall \ours yields the best mixing of clean and backdoored samples. Better mix means the model sees the clean and backdoored samples similarly, which also translates to low ASR.}
\label{fig:train-tsne-blendrs}
\end{figure*}

The proposed triggers are visualized in~\cref{fig:triggers} and the subsequently generated backdoored images are in~\cref{fig:vis-backdoor-app}. Training dynamics of \ours for a few more attacks along with the final image embedding projections via t-SNE can be found in Figures~\ref{fig:tsne-badnet-stripes} and~\ref{fig:train-tsne-blendrs}. t-SNE embedding comparison for the standard random noise based \badnet~\cite{gu2017badnets} is in~\cref{fig:train-tsne-badnet-random}.

\begin{table}[t]
\centering 
\begin{tabular}{c|c|c|c|c}
\toprule
& Poison & \ours & Re-Poison & \ours on Re-poison\\
\toprule
clean & 58.6 & 53.5 & 53.7 & 48.2\\
ASR & 98.6 & 0.1 & 98.0 & 0.1\\
\bottomrule
\end{tabular}
\vspace{-2mm}
\caption{\textbf{Adaptive attack on \ours.} \ours cleans the re-poisoned model perfectly, despite  the original backdoored model being a feasible solution.}
\label{tab:repoison}
\end{table}

\begin{table}[t]
\centering
\small

%\vspace{2mm}
\tabcolsep=1.1pt
\extrarowheight=1.25pt
\newl=9mm
\newlc=12mm
\begin{tabular}{L{18mm} *{2}{|C{\newlc} C{\newlc}}}
\toprule
& \multicolumn{4}{c}{COCO Precision@Top-1} \\
& \multicolumn{2}{c}{Image ret.} & \multicolumn{2}{c}{Text ret.} \\
\cline{2-5} 
Method & clean & bkd ($\uparrow$) & clean & bkd ($\uparrow$) \\
\toprule
\addlinespace[0.8mm]
\multicolumn{5}{l}{\textbf{Attack:} \badnet~\cite{gu2017badnets}} \\
\midrule
\rowcolor{NewGray} \clip  & 33.0 & 4.0 & 47.9 & 1.1 \\
\cleanclip  & 31.6 & 23.2 & 44.3 & 32.5 \\
\ours  & 29.8 & \textbf{26.4} & 44.4 & \textbf{37.4} \\
\toprule
\addlinespace[0.8mm]
\multicolumn{5}{l}{\textbf{Attack:} \badnet-Stripes}\\
\midrule
\rowcolor{NewGray}\clip  & 33.1 & 7.4 & 48.0 & 0.3 \\
\cleanclip  & 31.6 & 15.5 & 43.8 & 14.8 \\
\ours & 30.3 & \textbf{18.0} & 44.7 & \textbf{21.8} \\
\toprule
\addlinespace[0.8mm]
\multicolumn{5}{l}{\textbf{Attack:} \blended~\cite{chen2017targeted}}\\
\midrule
\rowcolor{NewGray}\clip  & 32.8 & 6.0 & 47.7 & 0.5\\
\cleanclip  & 31.8 & 17.1 & 44.3 & 18.4\\
\ours & 30.9 & \textbf{25.8} & 44.9 & \textbf{39.6}\\
\toprule 
\addlinespace[0.8mm]
\multicolumn{5}{l}{\textbf{Attack:} \blended-Stripes}\\
\midrule
\rowcolor{NewGray}\clip & 32.8 & 3.7 & 46.9 & 0.9\\
\cleanclip  & 31.6 & 9.6 & 44.5 & 8.6\\
\ours  & 30.5 & \textbf{18.9 }& 44.0 & \textbf{27.9 }\\
\toprule 
% \addlinespace[0.8mm]
% \multicolumn{5}{l}{\textbf{Attack:} SIG~\cite{barni2019new}}\\
% \midrule
% \clip & 32.1 & 7.8 & 46.6 & 3.9\\
% \cleanclip  & 31.1 & 13.9 & 43.6 & 15.8 \\
% \ours  & 30.5 & 18.9 & 44.0 & 27.9 \\
% \toprule
\addlinespace[0.8mm]
\multicolumn{5}{l}{\textbf{Attack:} \wanet~\cite{nguyen2021wanet}}\\
\midrule
\rowcolor{NewGray}\clip  & 32.6 & 2.5 & 46.9 & 0.0\\
\cleanclip  &31.1 & \textbf{18.2} & 43.2 & 23.1 \\
\ours  & 30.8 & 16.3 & 45.3 & \textbf{24.7}\\
\toprule
\addlinespace[0.8mm]
\multicolumn{5}{l}{\textbf{Attack:} \badclip~\cite{liang2024badclip}}\\
\midrule
\rowcolor{NewGray}\clip  & 32.6 & 2.3 & 48.1 & 0.4 \\
\cleanclip  & 31.3& 15.4 & 44.7 & 16.3\\
\ours & 30.1 & \textbf{20.1} & 45.3 & \textbf{23.2}\\

\bottomrule
 
\end{tabular}
\caption{\textbf{COCO retrieval for ResNet50 based \clip.} In compliment to~\cref{tab:rn50-novel}, we report Precision@Top-1 for both image-to-text and text-to-image retrieval for the COCO validation set. ``bkd'' shows when the inputs are with the backdoor trigger. The original poisoned \clip model is \colorbox{NewGray}{highlighted}.} 
\label{tab:rn50-cocoretrieval}
\end{table}

\begin{table}[t]
\centering
\small

%\vspace{2mm}
\tabcolsep=1.2pt
\extrarowheight=1.25pt
\newl=9mm
\newlc=12mm
\begin{tabular}{L{18mm} *{2}{|C{\newlc} C{\newlc}}}
\toprule
& \multicolumn{4}{c}{COCO Precision@Top-1} \\
& \multicolumn{2}{c}{Image ret.} & \multicolumn{2}{c}{Text ret.} \\
\cline{2-5} 
Method & clean & bkd ($\uparrow$) & clean & bkd ($\uparrow$) \\
\toprule
\addlinespace[0.8mm]
\multicolumn{5}{l}{\textbf{Attack:} \badnet-Stripes}\\
\midrule
\rowcolor{NewGray}\clip  & 33.5 & 1.7 & 47.8 & 1.1\\
\cleanclip  & 32.0 &7.6 & 45.3 & 5.4 \\
\ours  & 30.8 & \textbf{13.7} & 44.6 & \textbf{17.6} \\
\toprule
\addlinespace[0.8mm]
\multicolumn{5}{l}{\textbf{Attack:} \blended-Stripes}\\
\midrule
\rowcolor{NewGray}\clip  & 33.3 & 5.9 & 47.4 & 0.1 \\
\cleanclip  & 32.4 & 21.6 &45.9 & 27.5\\
\ours  & 30.5 & \textbf{26.1} & 39.9 & \textbf{36.6} \\
\toprule 
\addlinespace[0.8mm]
\multicolumn{5}{l}{\textbf{Attack:} \blended-Text}\\
\midrule
\rowcolor{NewGray}\clip  & 32.8 & 4.3 & 45.5 & 0.1 \\
\cleanclip & 31.7 & \textbf{15.1} & 44.4 & 14.0 \\
\ours & 31.8 & 12.5 & 45.2 & \textbf{18.3} \\
\toprule 
\addlinespace
\multicolumn{5}{l}{\textbf{Attack:} \blended-Triangles}\\
\midrule
\rowcolor{NewGray}\clip  & 32.9 & 3.1 & 47.0 & 0.0 \\
\cleanclip & 32.2 & 2.5 & 44.6 & 0.6\\
\ours  & 31.1 & \textbf{3.2} & 43.4 & \textbf{2.1} \\
\toprule 
\addlinespace[0.8mm]
\multicolumn{5}{l}{\textbf{Attack:} \badclip~\cite{liang2024badclip}}\\
\midrule
\rowcolor{NewGray}\clip & 33.5 & 2.3 & 48.0 & 0.0 \\
\cleanclip  & 32.1 & \textbf{18.9} &45.6 & \textbf{20.8}  \\
\ours  & 31.1 & 15.6 & 43.6 & 16.6\\
\bottomrule
 
\end{tabular}
\caption{\textbf{COCO retrieval for \vit-B/32 based \clip.} In compliment to~\cref{tab:vitb-novel}, we report Precision@Top-1 for both image-to-text and text-to-image retrieval for the COCO validation set. ``bkd'' shows when the inputs are with the backdoor trigger. The original poisoned \clip model is \colorbox{NewGray}{highlighted}.} 
\label{tab:vitb-cocoretrieval}
\end{table}

\end{document}